\documentclass[journal,twocolumn]{IEEEtran}

\usepackage{amsmath}
\usepackage{amssymb,latexsym}
\usepackage{amsfonts}
\usepackage{amscd}
\usepackage{algorithm}
\usepackage{algorithmic}
\usepackage{multirow}
\usepackage{changepage}
\usepackage{times}
\usepackage{float} 
\usepackage{graphics,subfigure}
\usepackage[latin1]{inputenc} 
 \usepackage{graphicx}
\usepackage{color}

\newtheorem{theorem}{\bf{Theorem}}

\graphicspath{{figures/}}

\newcommand{\cb}[1]{{\boldsymbol{#1}}}
\newcommand{\cp}[1]{\ifmmode {\mathcal{#1}}\else ${\mathcal{#1}}$\fi}
\newcommand{\bs}{\boldsymbol}

\def\tr{\text{trace}}
\def\Jex{J_\text{ex}}
\def\Jmin{J_{\min}}
\def\Jmse{J_\text{ms}}

\newcommand{\rmd}{r_{\text{md}}}
\newcommand{\rod}{r_{\text{od}}}
\newcommand{\bd}{\boldsymbol{d}}
\newcommand{\bfun}{\boldsymbol{f}}
\newcommand{\bI}{\boldsymbol{I}}
\newcommand{\bK}{\boldsymbol{K}}

\newcommand{\T}{{\bs T}}
\newcommand{\vp}{{\bs p}}
\newcommand{\vu}{\boldsymbol{u}}
\newcommand{\vun}{\boldsymbol{u}_n}
\newcommand{\mC}{{\bs C}}

\newcommand{\mG}{{\bs G}}
\newcommand{\mR}{{\bs R}}
\newcommand{\mI}{{\bs I}}
\newcommand{\vc}{{\bs c}}
\newcommand{\vy}{{\bs y}}
\newcommand{\bv}{\boldsymbol{v}}
\newcommand{\vv}{\bv_n}
\newcommand{\vk}{\bs {\kappa}_{\omega,n}}

\newcommand{\mO}{{\bs O}}
\newcommand{\mQ}{{\bs Q}}
\newcommand{\mRuu}{{\bs R}_{\vu\vu}}
\newcommand{\mRkk}{{\bs R}_{\kappa\kappa}}

\newcommand{\balpha}{\boldsymbol{\alpha}}
\newcommand{\bkappa}{\boldsymbol{\kappa}}
\newcommand{\sbt}{\,\begin{picture}(-1,1)(-1,-3)\circle*{3}\end{picture}\ }

\def\R{\ensuremath{{\mathrm{I\!R}}}}

\begin{document}

\title{Online dictionary learning for kernel LMS \\ Analysis and forward-backward splitting algorithm}

\author{Wei Gao, \textit{Student Member, IEEE,}
        	    Jie Chen, \textit{Student Member, IEEE} \\
        	    C\'edric~Richard, \textit{Senior Member, IEEE,}
        	    Jianguo Huang, \textit{Senior Member, IEEE}
\IEEEcompsocitemizethanks{This work was partially supported by the National Natural Science Foundation of China (61271415).}
\thanks{Wei Gao is with the Universit\'e de Nice Sophia-Antipolis, CNRS, Observatoire de la C\^ote d'Azur, France and the College of Marine Engineering, Northwestern Polytechnical University, China (gao\_wei@mail.nwpu.edu.cn)}
\thanks{Jie Chen and C\'edric~Richard are with the Universit\'e de Nice Sophia-Antipolis, CNRS, Observatoire de la C\^ote d'Azur, France (jie.chen@unice.fr; cedric.richard@unice.fr)}
\thanks{Jianguo Huang is with the College of Marine Engineering, Northwestern Polytechnical University, China (jghuang@nwpu.edu.cn)}
} 
 
\maketitle


\begin{abstract}

Adaptive filtering algorithms operating in reproducing kernel Hilbert spaces have demonstrated superiority over their linear counterpart for nonlinear system identification. Unfortunately, an undesirable characteristic of these methods is that the order of the filters grows linearly with the number of input data. This dramatically increases the computational burden and memory requirement. A variety of strategies based on dictionary learning have been proposed to overcome this severe drawback. Few, if any, of these works analyze the problem of updating the dictionary in a time-varying environment. In this paper, we present an analytical study of the convergence behavior of the Gaussian least-mean-square algorithm in the case where the statistics of the dictionary elements only partially match the statistics of the input data. This allows us to emphasize the need for updating the dictionary in an online way, by discarding the obsolete elements and adding appropriate ones. We introduce a kernel least-mean-square algorithm with $\ell_1$-norm regularization to automatically perform this task. The stability in the mean of this method is analyzed, and its performance is tested with experiments.
\end{abstract}


\section{Introduction}
\label{sec:intro}

Recently, adaptive filtering in reproducing kernel Hilbert spaces (RKHS) has become an appealing tool in many practical fields, including biomedical engineering, remote sensing and control. This framework allows the use of linear algorithms in the parameters for nonlinear system identification. It consists of mapping the original input data into a RKHS, and applying a linear adaptive filtering technique to the resulting functional data.

\begin{figure}[t]
	\centering\includegraphics[scale=0.5]{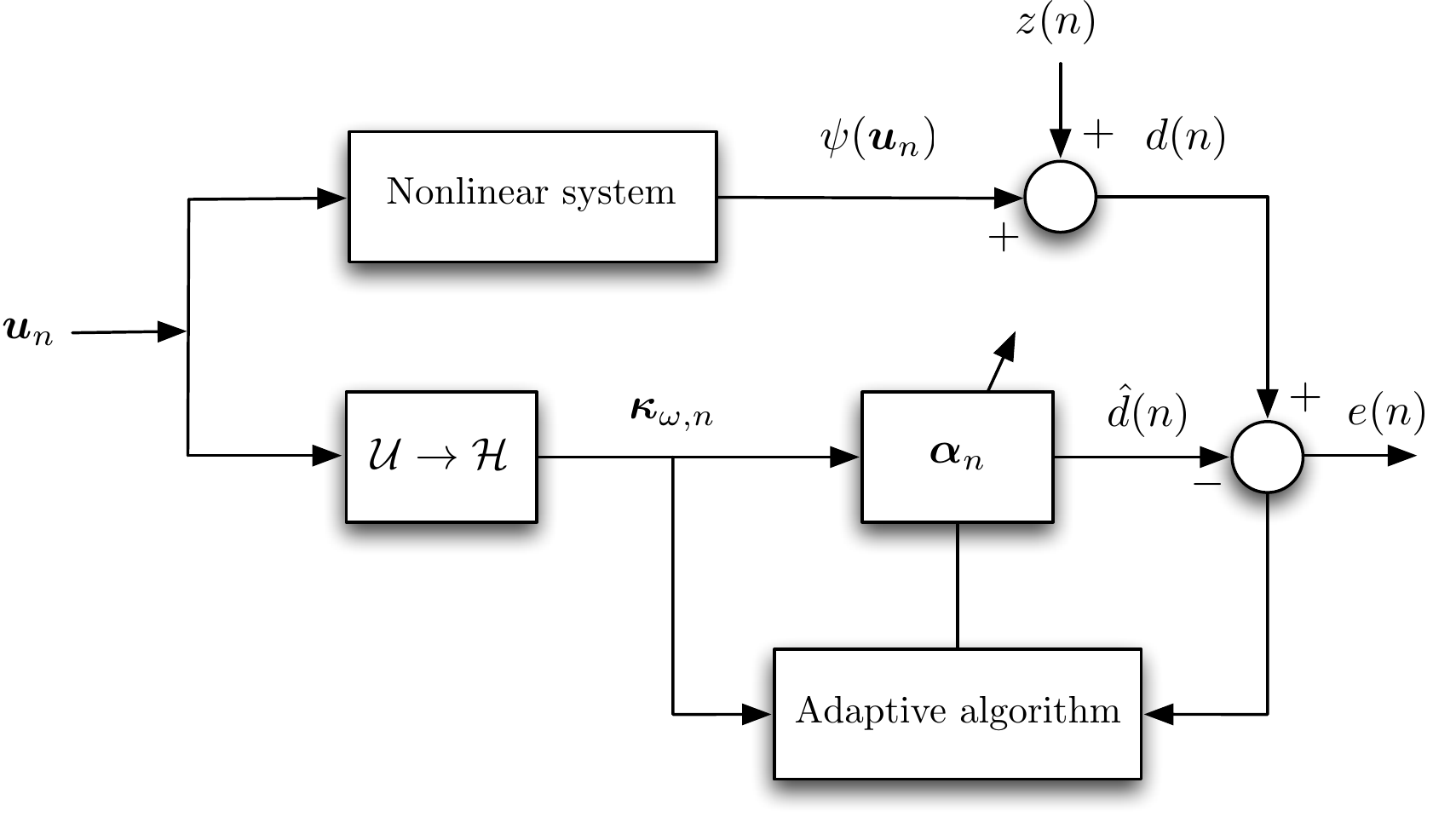}
	\caption{Kernel-based adaptive system identification.}
	\label{fig:identification.problem}
\end{figure}

The block diagram presented in Figure \ref{fig:identification.problem} presents the basic principles of this strategy. The subspace $\cp{U}$ is a compact of $\R^q$, $\kappa\!:\cp{U} \times \cp{U}\rightarrow\R$ is a reproducing kernel, and $(\cp{H},\langle\cdot,\!\cdot\rangle_\cp{H})$ is the induced RKHS with its inner product. Usual kernels involve, e.g., the radially Gaussian and Laplacian kernels, and the $q$-th degree polynomial kernel. The additive noise $z(n)$ is supposed to be white and zero-mean, with variance $\sigma_z^2$. Considering the least-squares approach, given $N$ input vectors $\vun$ and desired outputs $d_n$, the identification problem consists of determining the optimum function $\psi^*(\cdot)$ in $\cp{H}$ that solves the problem
\begin{equation}
\vspace{-1.5mm}
    	\label{eq:problem.functional}
    	\psi^*=\mathop{\arg\min}_{\psi\in\cp{H}} \Big\{J(\psi)=\sum_{i=1}^N(d_i-\psi(\cb{u}_i))^2 \Big\}.
\end{equation}
By virtue of the representer theorem~\cite{Sch00}, the function $\psi(\cdot)$ can be written as a kernel expansion in terms of available training data, namely, $\psi(\cdot)=\sum_{j=1}^N\alpha_j\,\kappa(\cdot,\cb{u}_j)$. The above optimization problem becomes
\begin{equation}
\vspace{-1.5mm}
    	\label{eq:problem.parametric}
    	\balpha^*=\mathop{\arg\min}_{\balpha\in\R^N}\Big\{J(\balpha)=\sum_{j=1}^N(d_j-\balpha^\top\bkappa_j)^2\Big\}.
\end{equation}
where $\bkappa_j$ is the $(N \times 1)$ vector with $i$-th entry $\kappa(\cb{u}_i,\cb{u}_j)$. Online processing of time series data raises the question of
how to process an increasing amount $N$ of observations as new data is collected. Indeed, an undesirable characteristic of problem \eqref{eq:problem.functional}-\eqref{eq:problem.parametric} is that the order of the filters grows linearly with the number of input data. This dramatically increases the computational burden and memory requirement. To overcome this drawback, several authors have focused on fixed-size models of the form
\begin{equation}
          \vspace{-1.5mm}
    	\label{eq:model.expansion}
    	\psi(\cdot)=\sum_{j=1}^M\alpha_j\,\kappa(\cdot,\cb{u}_{\omega_j}).
\end{equation}
We call $\cp{D}=\{\kappa(\cdot,\cb{u}_{\omega_j})\}_{j=1}^M$ the dictionary, which has to be learnt from input data, and $M$ the order of the kernel expansion by analogy with linear transversal filters. Online identification of kernel-based models generally relies on a two-step process at each iteration: a model order control step that updates the dictionary, and a parameter update step. This two-step process is the essence of most adaptive filtering techniques with kernels.

Based on this scheme, several state-of-the-art linear methods were reconsidered to derive powerful nonlinear generalizations operating in high-dimensional RKHS \cite{Sayed2003, Haykin1991}: the recursive least-squares algorithm (RLS), the affine projection algorithm (APA), and the least-mean-square algorithm (LMS). On the one hand, the kernel recursive least-squares algorithm was introduced in \cite{Engel2004}. The sliding-window KRLS and and extended KRLS algorithms were successively derived in \cite{Vaerenbergh2006, Liu2009a}. More recently, the KRLS tracker algorithm was introduced in \cite{Vaerenbergh2012}, with ability to forget past information using forgetting strategies. This allows the algorithm to track non-stationary input signals based on the idea of the exponentially-weighted KRLS algorithm \cite{Liu2010}. On the other hand, the kernel affine projection algorithm (KAPA) and, as a particular case, the kernel normalized LMS algorithm (KNLMS), were independently introduced in \cite{honeine2007line,Richard2009,Slavakis2008,Liu2008b}. The kernel least-mean-square algorithm (KLMS) was presented in \cite{richard2005filtrage,Liu2008a}, and attracted the attention because of its simplicity and robustness. A very detailed analysis of the stochastic behavior of the KLMS algorithm with Gaussian kernel was provided in \cite{Parreira2012}, and a closed-form condition for convergence was recently introduced in \cite{richard2012closed}. The quantized KLMS algorithm (QKLMS) was proposed in \cite{Chen2012a}, and the QKLMS algorithm with $\ell_1$-norm regularization was introduced in \cite{Chen2012b}. Note that the latter uses $\ell_1$-norm in order to sparsify the parameter vector $\balpha$ in the kernel expansion \eqref{eq:model.expansion}. A subgradient approach was considered to accomplish this task, which contrasts with the more efficient forward-backward splitting algorithm recommended in \cite{Yukawa2012,gao2013kernel}. A recent trend within the area of adaptive filtering with kernels consists of extending all the algorithms to give them the ability to process complex input signals \cite{Bouboulis2011, Bouboulis2012}. The convergence analysis of the complex KLMS algorithm with Gaussian kernel presented in \cite{Paul2012} is a direct application of the derivations in \cite{Parreira2012}. Finally, quaternion kernel least-squares algorithm was recently introduced in \cite{Tobar2013}.

All the above-mentioned methods use more or less sophisticated learning strategies to decide, at each time instant $n$, whether $\kappa(\cdot,\cb{u}_n)$ deserves to be inserted into the dictionary or not. One of the most informative criteria uses approximate linear dependency (ALD) condition to test the ability of the dictionary elements to approximate the current input $\kappa(\cdot,\cb{u}_n)$ linearly \cite{Engel2004}. Other well-known criteria include the novelty criterion \cite{Platt1991}, the coherence criterion \cite{Richard2009}, the surprise criterion~\cite{Liu2009b}, and closed-ball sparsification criterion \cite{Slavakis2013}. Without loss of generality, we focus on KLMS algorithm with coherence sparsification (CS) due to its simplicity and effectiveness. Most of these strategies for dictionary update are only able to incorporate new elements into the dictionary, and to possibly forget the old ones using a forgetting factor. This means that they cannot automatically discard obsolete kernel functions, which may be a severe drawback within the context of a time-varying environment. Recently, sparsity-promoting regularization was considered within the context of linear adaptive filtering. All these works propose to use, either the $\ell_1$-norm of the vector of filter coefficients as a regularization term, or some other related regularizers to limit the induced bias. The optimization procedures consist of subgradient descent \cite{Chen2009}, projection onto the $\ell_1$-ball \cite{Slavakis2010}, or online forward-backward splitting \cite{Murakami2010}. Surprisingly, this idea was little used within the context of kernel-based adaptive filtering. To the best of our knowledge, only \cite{Yukawa2012} uses projection for least-squares minimization with weighted block $\ell_1$-norm regularization, within the context of multi-kernel adaptive
filtering.

Few, if any, of these works strictly analyze the necessity of updating the dictionary in a time-varying environment. In this paper, we present an analytical study of the convergence behavior of the Gaussian least-mean-square algorithm in the case where the statistics of the dictionary elements only partially match the statistics of the input data. This allows us to emphasize the need for updating the dictionary in an online way, by discarding the obsolete elements and adding appropriate ones. Then, we introduce a KLMS algorithm with $\ell_1$-norm regularization to automatically perform this task. The stability in the mean of this method is analyzed, and its performance is tested with experiments.

\section{Behavior analysis of Gaussian KLMS algorithm with partially matching dictionary}
\label{sec:BeahviorAnalysis}

We shall now extend the analysis of the Gaussian KLMS algorithm depicted in \cite{Parreira2012} in the case where a given proportion of the dictionary elements has distinct stochastic properties from the input samples. This will allows us to justify the need for updating the dictionary in an online way. It is interesting to note that the following analysis was made possible by greatly simplifying the derivations in \cite{Parreira2012}. This simplification should allow us to analyze other variants of the Gaussian KLMS algorithm, including the multi-kernel case, in future research works.

\subsection{KLMS algorithms}
\label{subsec:MSEanalysis}

Several versions of the KLMS algorithm have been proposed in the literature. The two most significant versions
consist of considering the problem \eqref{eq:problem.functional} and performing gradient descent on the function $\psi(\cdot)$ in $\cp{H}$, or considering
the problem~\eqref{eq:problem.parametric} and performing gradient descent on the parameter vector $\balpha$, respectively. The former strategy is
considered in \cite{Liu2008a} for instance, while the latter is applied in \cite{Richard2009}. Both need the use of an extra mechanism for
controlling the order $M$ of the kernel expansion \eqref{eq:model.expansion} at each time instant $n$. We shall now introduce such a model
order selection stage, before briefly introducing the parameter update stage we recommend.

\subsubsection{Dictionary update}

Coherence is a fundamental parameter to characterize a dictionary in linear sparse approximation problems. It was redefined in \cite{Richard2009} within the context of adaptive filtering with kernels as follows:
\begin{equation}
    	\label{eq:coherence}
	\mu=\max_{i \neq j}|\kappa(\cb{u}_{\omega_i},\cb{u}_{\omega_j})|
\end{equation}
where $\kappa$ is a unit-norm kernel. Coherence criterion suggests inserting the candidate input $\kappa(\cdot,\vu_n)$ into the dictionary provided that its coherence remains below a given threshold~$\mu_0$
\begin{equation}
    	\label{eq:coherence.rule}
    	\max_{m=1,\ldots,M}|\kappa(\cb{u}_n,\cb{u}_{\omega_m})|\leq\mu_0,
\end{equation}
where $\mu_0$ is a parameter in $[0,1[$ determining both the level of sparsity and the coherence of the dictionary. Note that the quantization criterion introduced in \cite{Chen2012a} consists of comparing $\min_{m=1,\ldots,M}\|\cb{u}_n-\cb{u}_{\omega_m}\|$ with a threshold. It is thus strictly equivalent to the original coherence criterion in the case of radially kernels such as the Gaussian one considered hereafter.

\subsubsection{Filter parameter update}
\label{subsubsec:Filterparat}

At iteration $n$, upon the arrival of new data $(\vu_n,d_n)$, one of the following alternatives holds. If $\kappa(\cdot,\cb{u}_{n})$ does not satisfy the coherence rule \eqref{eq:coherence.rule}, the dictionary remains unaltered. On the other hand, if condition \eqref{eq:coherence.rule} is met, kernel function $\kappa(\cdot, \cb{u}_{n})$ is inserted into the dictionary where it is then denoted by $\kappa(\cdot,\cb{u}_{\omega_{M+1}})$. The least-mean-square algorithm applied to the parametric form \eqref{eq:problem.parametric} leads to the following algorithm \cite{Richard2009}
\begin{itemize}
	\item Case 1: $\max_{m=1,\ldots,M}|\kappa(\cb{u}_{n},\cb{u}_{\omega_m})|>\mu_0$
	\begin{equation}
		\label{eq:klms.perso.1}
		\balpha_n = \balpha_{n-1}+\eta\,e_n\,\bkappa_{\omega,n}
	\end{equation}
	\item Case 2: $\max_{m=1,\ldots,M}|\kappa(\cb{u}_{n},\cb{u}_{\omega_m})|\leq\mu_0$
	\begin{equation}
		\label{eq:klms.perso.2}
		\balpha_n=\left(\!\begin{array}{ccc}{\balpha_{n-1}} \\
		{0}\end{array}\!\right)+\eta\,e_n\,\bkappa_{\omega,n}
	\end{equation}
\end{itemize}
where $\vk=[\kappa(\vun,\vu_{\omega_1}),\cdots,\kappa(\vun,\vu_{\omega_M})]^\top$.

The coherence criterion guarantees that the dictionary dimension is finite for any input sequence $\{{\vu_n}\}_{n=1}^{\infty}$ due to the compactness of the input space $\mathcal U$ in \cite[proposition 2]{Richard2009}.

The KLMS algorithm derived in \cite{Chen2012a} adopts the Fr\'echet's notion of differentiability to derive a gradient descent direction with respect to $\psi(\cdot)$ in problem \eqref{eq:problem.functional}, that is,
\begin{equation}
	\label{eq:frechet.dif}
	\nabla E\{(d_n-\psi(\vu_n))^2\} = -2\,E\{e_n\,\kappa(\cdot,\vu_n)\} \approx -2\,e_n\,\kappa(\cdot,\vu_n)
\end{equation}
A consequence of equation \eqref{eq:frechet.dif} is that this principle leads to the update of one component of $\balpha_n$ at each iteration. The resulting algorithm reduces to a coordinate stochastic gradient descent. In the following, we shall focus on parameter update rules \eqref{eq:klms.perso.1}-\eqref{eq:klms.perso.2}.

\subsection{Mean square error analysis}

Consider the nonlinear system identification problem {shown in} Figure~\ref{fig:identification.problem}, and the finite-order model \eqref{eq:model.expansion} based on the Gaussian kernel
\begin{equation}
	\label{eq:gaussian.kernel}
	\kappa(\vu_i,\vu_j)=\exp\left(\frac{-\| \vu_i - \vu_j \|^2}{2\xi^2}\right)
\end{equation}
where $\xi$ is the kernel bandwidth. The nonlinear system input data $\vun\in \R^{q\times 1}$ are supposed to be zero-mean, independent, and identically distributed Gaussian vector. We consider that the entries of $\vun$ can be correlated, and we denote by $\mRuu=E\{\vun\vun^\top\}$ the autocorrelation matrix of the input data. It is assumed that the input data $\vun$ or the transformed inputs by kernel $\psi(\vun)$ are locally or temporally stationary in the environment needed to be analyzed. The estimated system output is given by
\begin{equation}
	\label{eq:desired.estimator}
	\hat{d}_n= \balpha_n^{\top}\,\vk
\end{equation}
with $\balpha_n=[\alpha_1(n), \ldots, \alpha_M(n)]^{\top}$.  The corresponding estimation error is defined as
\begin{equation}
	\label{eq:error1}
	e_n=d_n-\hat{d}_n.
\end{equation}
Squaring both sides of \eqref{eq:error1} and taking the expected value leads to the mean square error (MSE)
\begin{equation}
	\label{eq:normaleq}
	\Jmse(n)=E\{ e^2_n \} = E\{d^2_n\} - 2 \vp^{\top}_{\kappa d}\,\balpha_n + \balpha^{\top}_n\,\mRkk\,\balpha_n
\end{equation}
where $\mRkk=E\{\vk\vk^{\top}\}$ is the correlation matrix of the kernelized input $\vk$, and $\vp_{\kappa d} = E\{d_n\,\vk\}$ is the cross-correlation vector between $\vk$ and $d_n$. It has already been proved that $\mRkk$ is strictly positive definite \cite{Parreira2012}. Thus, the optimum weight vector is given by
\begin{equation}
	\label{eq:wienereq}
	{\balpha}_{\text{opt}} = \mRkk^{-1}\,\vp_{\kappa d}
\end{equation}
and the corresponding minimum MSE is
\begin{equation}
	\label{eq:minerror}
	J_{\min} = E\{d^2_n\} - \vp_{\kappa d}^{\top}\,\mRkk^{-1}\,\vp_{\kappa d}.
\end{equation}
Note that expressions of \eqref{eq:wienereq} and \eqref{eq:minerror} are the well-known Wiener solution and minimum MSE, respectively, where
the input signal vector has been replaced by the kernelized input vector.

In order to determine $\balpha_{\text{opt}}$, we shall now calculate the correlation matrice $\mRkk$ using the statistical properties of the input $\vun$ and the kernel definition. Let us introduce the following notations
\begin{equation}
	\|\vun-\vu_{\omega_i}\|^2 + \| \vun-\vu_{\omega_j}\|^2=\vy^{\top}_3\,\mQ_3\,\vy_3
\end{equation}
where $\|\cdot\|$ is the $\ell_2$-norm and
\begin{equation} \label{eq:vectorsy}
	\begin{split}
	\vy_3=\left(\vun^{\top}\ \vu^{\top}_{\omega_i}\ \vu^{\top}_{\omega_j}\right)^{\top}
	\end{split}
\end{equation}
 and
\begin{equation} \label{eq:matricesQ}
	\mQ_{3} = \left( \begin{array}{ccc}
		2\mI & -\mI & -\mI\\
		-\mI & \mI & \mO \\
		-\mI & \mO &\mI
		\end{array}\right)
\end{equation}
where $\mI$ is the $(q\times q)$ identity matrix, and $\mO$ is the $(q\times q)$ null matrix. From \cite[p. 100]{Omura1965}, we know that the moment generating function of a quadratic form $z=\vy^{\top}\mQ\,\vy$, where $\vy$ is a zero-mean Gaussian vector with covariance $\mR_{y}$, is given~by
\begin{equation}
	\label{eq:omura}
	\psi_z(s) = E \{ e^{sz} \} = \det\{\mI - 2s\mQ\mR_y\}^{-1/2}.
\end{equation}
Making $s=-1/(2\xi^2)$ in equation \eqref{eq:omura}, we find that the $(i,j)$-th element of $\mRkk$ is given by
\begin{equation}
	\label{eq:Rkk_mdf}
	\left[\mRkk\right]_{ij} =
	\begin{cases}
		\begin{split}
			\rmd&=\det\left\{\mI_3 + \mQ_{3}\,\mR_{3}(i,j)/\xi^2\right\}^{-1/2}, &i=j \\
			\rod&=\det\left\{\mI_3 + \mQ_{3}\,\mR_{3}(i,j)/\xi^2\right\}^{-1/2}, &i\neq j
		\end{split}
	\end{cases}
\end{equation}
with $1\leq i,\, j \leq M$, and $\rmd$ and $\rod$ are the main-diagonal and off-diagonal entries of $\mRkk$, respectively. In equation~\eqref{eq:Rkk_mdf}, $\mR_{\ell}$ is the $(\ell q\times \ell q)$ correlation matrix of vector $\vy_{\ell}$, $\mI_{\ell}$ is the $(\ell q\times \ell q)$ identity matrix, and $\det\{\cdot\}$ denotes the determinant of a matrix. Cases $(i=j)$ and $(i\neq j)$ correspond to different forms of $\mR_{3}(i,j)$, given by
\begin{equation}
	\label{eq:R3ij}
           \mR_{3}(i,j) = 
           \left( \begin{array}{ccc}
		                \mR_{\vu\vu} 	& \mO 			& \mO \\
	   	                \mO 			& \mR_\cp{D}(i,i) 	& \mR_\cp{D}(i,j) \\
		               \mO 			& \mR_\cp{D}(i,j)	&\mR_\cp{D}(j,j)
		\end{array}\right)
\end{equation}
where $\mR_\cp{D}(i,j)=E\{\vu_{\omega_i}\vu^{\top}_{\omega_j}\}$ is the intercorrelation matrix of the dictionary elements. Compared with \cite{Parreira2012}, the formulations \eqref{eq:Rkk_mdf}-\eqref{eq:R3ij}, and other reformulations pointed out in the following, allow to address more general problems by making the analyses tractable. In particular, in order to evaluate the effects of a mismatch between the input data and the dictionary elements, we shall now consider the case where that they do not necessarily share the same statistical properties.

Suppose now that the first $L$ dictionary elements $\{ \vu_{\omega_m} \in \R^q: 1\leq m \leq L\}$ have the same autocorrelation matrix $\mR_{\vu\vu}$ as the input $\vun$, whereas the other $(M-L)$ elements $\{\vu_{\omega_m} \in \R^q: L< m \leq M\}$ have a distinct autocorrelation matrix denoted by $\tilde{\mR}_{\vu\vu}$. Such a situation may occur in a time-varying environment with most, if not all, of the existing strategies for dictionary update: they are only able to incorporate new elements into the dictionary, and cannot automatically discard obsolete kernel functions. In this case, $\mR_\cp{D}(i,j)$ in equation~\eqref{eq:R3ij} writes
\begin{equation}
          \label{eq:def.blk}
          \mR_\cp{D}(i,j) = \begin{cases} \mR_{\vu\vu} , \quad 1\leq i=j \leq L\\
          								\widetilde{\mR}_{\vu\vu} , \quad L < i=j \leq M\\
									\mO,  \qquad 1 \leq i \neq j \leq M
								       \end{cases}
\end{equation}
which allows to calculate the correlation matrix $\mRkk$ of the kernelized input via equation \eqref{eq:Rkk_mdf}. Note that $\mR_\cp{D}(i,j)$
in equation \eqref{eq:R3ij} reduces to $\delta_{ij}\,\mRuu$, with $\delta_{ij} = 1$ if $(i = j)$, otherwise $0$, in the case $(L = M)$ considered in \cite{Parreira2012}.

\subsection{Transient behavior analysis}
\label{subsec:Trasntbehavior}

\subsubsection{Mean weight behavior}

The weight update equation of KLMS algorithm is given by
\begin{equation}
	\label{eq:weightKLMS}
	\balpha_{n+1}=\balpha_n+\eta \, e_n \, \vk
\end{equation}
where $\eta$ is the step size. Defining the weight error vector $\bv_n=\balpha_n-\balpha_{\text{opt}}$ leads to the weight error vector update equation
\begin{equation}
	\label{eq:error_vector1}
	\bv_{n+1}=\vv + \eta\,e_n\,\vk.
\end{equation}
From \eqref{eq:desired.estimator} and \eqref{eq:error1}, and the definition of $\bv_n$, the error equation is given by
\begin{equation}
	\label{eq:error2}
	e_n = d_n -\vk^{\top}\,\vv - \vk^{\top}\,{\balpha}_{\text{opt}}
\end{equation}
and the optimal estimation error is
\begin{equation}
	\label{eq:min_error}
	e^o_n=d_n - \vk^{\top}\,\balpha_{\text{opt}}.
\end{equation}
Substituting \eqref{eq:error2} into \eqref{eq:error_vector1} yields
\begin{equation}
	\label{eq:error_vector2}
	\bv_{n+1}=\vv + \eta\,d_n\,\vk - \eta\,\vk^{\top}\,\vv\,\vk - \eta\,\vk^{\top}\,\balpha_{\text{opt}}\,\vk.
\end{equation}

Simplifying assumptions are required in order to make the study of the stochastic behavior of $\vk$ mathematically feasible. The so-called modified independence assumption (MIA) suggests that $\vk \vk^\top$ is statistically independent of~$\vv$. It is justified in detail in \cite{Minkoff2001}, and shown to be less restrictive than the independence assumption \cite{Sayed2003}. We also assume that the finite-order model provides a close enough approximation to the infinite-order model with minimum MSE, so that $E\{e^o_n\} \approx 0$. Taking the expected value of both sides of equation \eqref{eq:error_vector2} and using these two assumptions yields
\begin{equation}
	\label{eq:stab_mean}
	E\{\bv_{n+1}\} = (\mI-\eta\,\mRkk)\,E\{\bv_{n}\}
\end{equation}
This expression corresponds to the LMS mean weight behavior for the kernelized input vector $\vk$.

\subsubsection{Mean square error behavior}
\label{subsec:MSE}

Using equation \eqref{eq:error2} and the MIA, the second-order moments of the weights are related to the MSE through \cite{Sayed2003}
\begin{equation}
	\label{eq:defJms}
	J_\text{ms}(n)= \Jmin + \text{trace}\{\mRkk\mC_v(n)\}
\end{equation}
where $\mC_v(n)=E\{\vv\vv^{\top}\}$ is the autocorrelation matrix of the weight error vector $\vv$, $\Jmin=E\{{e^o_n}^2\}$ denotes the minimum MSE, and $\text{trace}\{\mRkk\mC_v(n)\}$ is the excess MSE (EMSE). The analysis of the MSE behavior \eqref{eq:defJms} requires a model for $\mC_v(n)$, which is highly affected by the kernelization of the input signal $\vun$. An analytical model for the behavior of $\mC_v(n)$ was derived in \cite{Parreira2012}. Using simplifying assumptions derived from the MIA, it reduces to the following recursion
\begin{subequations}
	\begin{equation}
		\label{eq:Cv}
		\begin{split}
		\mC_v(n+1) \approx \mC_v(n) &- \eta\,(\mRkk\mC_v(n) + \mC_v(n)\mRkk) \\
		 & + \eta^2\,\T(n) + \eta^{2}\,\mRkk\Jmin
		 \end{split}
	\end{equation}
\vspace{-3mm}
with
	\begin{equation}
	\vspace{-1mm}
		\label{eq:T}
		\T(n) = E\{\vk\,\vk^{\top}\,\vv\,\vv^{\top}\,\vk\,\vk^{\top}\}.
	\end{equation}
\end{subequations}
The evaluation of expectation \eqref{eq:T} is an important step in the analysis. It leads to extensive calculus if proceeding as in \cite{Parreira2012} because, as $\vk$ is a nonlinear transformation of a quadratic function of the Gaussian input vector $\vu_n$, it is neither zero-mean nor Gaussian. In this paper, we provide an equivalent approach that greatly simplifies the calculation. This allows us to consider the general case where there is possibly a mismatch between the statistics of the input data $\vu_n$ and the dictionary elements. Using the MIA to determine the $(i,j)$-th element of $\T(n)$ in equation \eqref{eq:T} yields
\begin{align}
	[\T(n)]_{ij} \approx\sum_{\ell=1}^{M} \sum_{p=1}^{M} E\{\kappa_{\omega,n}(i)\,\kappa_{\omega,n}(j)\,&\kappa_{\omega,n}(\ell)\,\kappa_{\omega,n}(p)\} \nonumber\\ 
	&\times [\mC_v(n)]_{\ell p}.     	\label{expT}
\end{align}
 where $\kappa_{\omega,n}(i)\!=\!\kappa(\vun,\vu_{\omega_i})$. This expression can be written~as
\begin{equation}
	\label{eq:expTK}
	[\T(n)]_{ij} \approx \tr \{\bK(i,j) \, \mC_v(n) \}
\end{equation}
where the $(\ell,p)$-th entry of $\bK(i,j)$ is given by $[\bK(i,j)]_{\ell,p} = E\{e^{sz}\}$, with $s=-1/(2\xi^2)$ and
\begin{equation}
	z = \|\vun-\vu_{\omega_i}\|^2 + \|\vun-\vu_{\omega_j}\|^2 + \|\vun-\vu_{\omega_\ell}\|^2 + \|\vun-\vu_{\omega_p}\|^2
\end{equation}
Using expression \eqref{eq:omura} leads us to
\begin{equation}
	\label{expKlp}
	[\bK(i,j)]_{\ell,p} = [\det\{\mI_5 + \mQ_{5}\,\mR_{5}(i,j,\ell,p)/\xi^2\}]^{-1/2}.
\end{equation}
with
\begin{equation}
		\mQ_{5} = \left( \begin{array}{ccccc}
 		4\mI& -\mI& -\mI& -\mI& -\mI \\
		-\mI& \mI& \mO & \mO & \mO \\
		-\mI& \mO & \mI& \mO & \mO \\
		-\mI& \mO & \mO & \mI& \mO \\
		-\mI& \mO & \mO & \mO & \mI
		\end{array}\right)
\end{equation}
and 
\begin{equation}
	\begin{split}
		&\mR_{5}(i,j,\ell,p) = \\
		&\left( \begin{array}{ccccc}
 		\mR_{\vu\vu}	& \mO			& \mO			& \mO				& \mO \\
		\mO			& \mR_\cp{D}(i,i)	& \mR_\cp{D}(i,j)	& \mR_\cp{D}(i,\ell)		& \mR_\cp{D}(i,p) 	\\
		\mO			& \mR_\cp{D}(i,j)	& \mR_\cp{D}(j,j)	& \mR_\cp{D}(j,\ell)		& \mR_\cp{D}(j,p) 	\\
		\mO			& \mR_\cp{D}(i,\ell)	& \mR_\cp{D}(j,\ell)	& \mR_\cp{D}(\ell,\ell)	& \mR_\cp{D}(\ell,p) 	\\
		\mO			& \mR_\cp{D}(i,p)	& \mR_\cp{D}(j,p)	& \mR_\cp{D}(\ell,p)		& \mR_\cp{D}(p,p) 	\\
		\end{array}\right),
	\end{split}
\end{equation}
which uses the same block definition as in \eqref{eq:def.blk}. Again, note that $\mR_\cp{D}(i,j)$ in the above equation reduces to $\delta_{ij}\,\mRuu$ in the regular case $(L = M)$ considered in \cite{Parreira2012}. This expression concludes the calculation.

\subsection{Steady-state behavior}

We shall now determine the steady-state of the recursion~\eqref{eq:Cv}. Observing that it only involves linear operations on the entries of $\mC_v(n)$, we can rewrite this equation in a vectorial form in order to simplify the derivations. The lexicographic representation of \eqref{eq:Cv} is as follows
\begin{equation}
	\label{eq:lex.C}
	\vc_v(n+1) = \mG\,\vc_v(n) + \eta^2 \Jmin\,{\bs r}_{\kappa\kappa}
	\vspace{-3mm}
\end{equation}
with
\begin{equation}
          \label{eq:vecCRkk}
           \mG = \mI - \eta(\mG_1+ \mG_2)+ \eta^2\mG_3
\end{equation}
where $\vc_v(n)$ and ${\bs r}_{\kappa\kappa}$ are the lexicographic representations of $\mC_v(n)$ and $\mRkk$, respectively. Matrix $\mG$ is found by the use of the following definitions: 

$\sbt$\, $\mI$ is the identity matrix of dimension $M^2 \times M^2$;

$\sbt$\, $\mG_1$ is involved in the product $\mC_v(n)\mRkk$. It is a block-diagonal matrix, with $\mRkk$ on its diagonal. It can thus be written as $\mG_1=\mI \otimes \mRkk$, where $\otimes$ denotes the Kronecker tensor product;

$\sbt$\, $\mG_2$ is involved in the product $\mRkk\mC_v(n)$, and can be written as $\mRkk \otimes \mI$;

 $\sbt$\, $\mG_3$ is the lexicographic representation of $\T(n)$ in equation~\eqref{eq:expTK}, namely,
\begin{equation}
                [{\mG_3}]_{i+(j-1)M,\ell+(p-1)M} =  [\bK(i,j)]_{\ell,p}
\end{equation}
with $1 \leq i,j,\ell,p \leq M$.

Note that $\mG_{1}$ to $\mG_{3}$ are symmetric matrices, which implies that $\mG$ is also symmetric. Assuming convergence, the closed-formed solution of the recursion \eqref{eq:lex.C} is given by
\begin{equation}
	\label{eq:cmd_vet2}
	\vc_v(n) = \mG^n\,\big[\vc_v(0) - \vc_v(\infty)\big] + \vc_v(\infty)
\end{equation}
where $\vc_v(\infty)$ denotes the vector $\vc_v(n)$ in steady-state, which is given by
\begin{equation}
	\label{eq:c_inf}
	\vc_v(\infty) = \eta^2\,\Jmin\,(\mI - \mG)^{-1}\,{\bs r}_{\kappa\kappa}
\end{equation}
From equation \eqref{eq:defJms}, the steady-state MSE is finally given by
\begin{equation}
	\label{eq:defJms_inf}
	J_\text{ms}(\infty) = \Jmin + \tr\{\mRkk\,\mC_v(\infty)\}
\end{equation}
where $\Jex(\infty)=\text{trace}\{\mRkk\,\mC_v(\infty)\}$ is the steady-state EMSE.

In the next section, simulation results will be provided to illustrate the validity of this model. This will allow us study of the convergence behavior of the algorithm in the case where the statistics of the dictionary elements only partially match the statistics of the input data.

\subsection{Simulation results}
\label{sec:AnalysisSimulation}

Two examples with abrupt variance changes in the input signal are presented hereafter. In each situation, the size of the dictionary was fixed beforehand, and the entries of the dictionary elements were i.i.d. randomly generated from a zero-mean Gaussian distribution. Each time series was divided into two subsequences. For the first one, the variance of this distribution was set as equal to the variance of the input signal. For the second one, it was abruptely set to a smaller or larger value in order to simulate a dictionary misadjustment. 

\noindent\underline{Notation}: In Tables \ref{tab:result_Exp1} and \ref{tab:result_Exp2}, dictionary settings are compactly expressed as $\cp{D}_i=\{M_i@\sigma_i\}\cup\{M_i'@\sigma'_i\}$. This has to be interpreted as: Dictionary $\cp{D}_i$ is composed of $M_i$ vectors with entries i.i.d. randomly generated from a zero-mean Gaussian distribution with standard deviation $\sigma_i$, and $M'_i$ vectors with entries i.i.d. randomly generated from a zero-mean Gaussian distribution with standard deviation $\sigma'_i$.

\subsubsection{Example 1}

Consider the problem studied in \cite{Parreira2012, Narendra1990, Mandic2004}, for which
\begin{eqnarray}
\begin{cases}
	y(n)= \displaystyle\frac{y(n-1)}{{1 + y^2(n-1)}} + u^3(n-1) \\
        d(n)= y(n) + z(n)
\end{cases}
\end{eqnarray}
where the output signal $y(n)$ was corrupted by a zero-mean i.i.d. Gaussian noise $z(n)$ with variance $\sigma_z^2=10^{-4}$. The input sequence $u(n)$ was i.i.d. randomly generated from a zero-mean Gaussian distribution with two possible standard deviations, $\sigma_u = 0.35$ or $0.15$, to simulate an abrupt change between two subsequences. The overall length of the input sequence was $4\times 10^4$. Distinct dictionaries, denoted by $\cp{D}_1$ and $\cp{D}_2$, were used for each subsequence. The Gaussian kernel bandwidth $\xi$ was set to $0.02$, and the KLMS step-size $\eta$ was set to $0.01$. Two situations were investigated. For the first one, the standard deviation of the input signal was changed from $0.35$ to $0.15$ at time instant $n=2\times 10^4$. Conversely, in the second one, it was changed from $0.15$ to $0.35$.

\begin{figure*}[!htb]
\subfigure[$\cp{D}_2=\{10@0.35\}$]
{\begin{minipage}[b]{0.333\textwidth}
	\includegraphics[width=5.5cm]{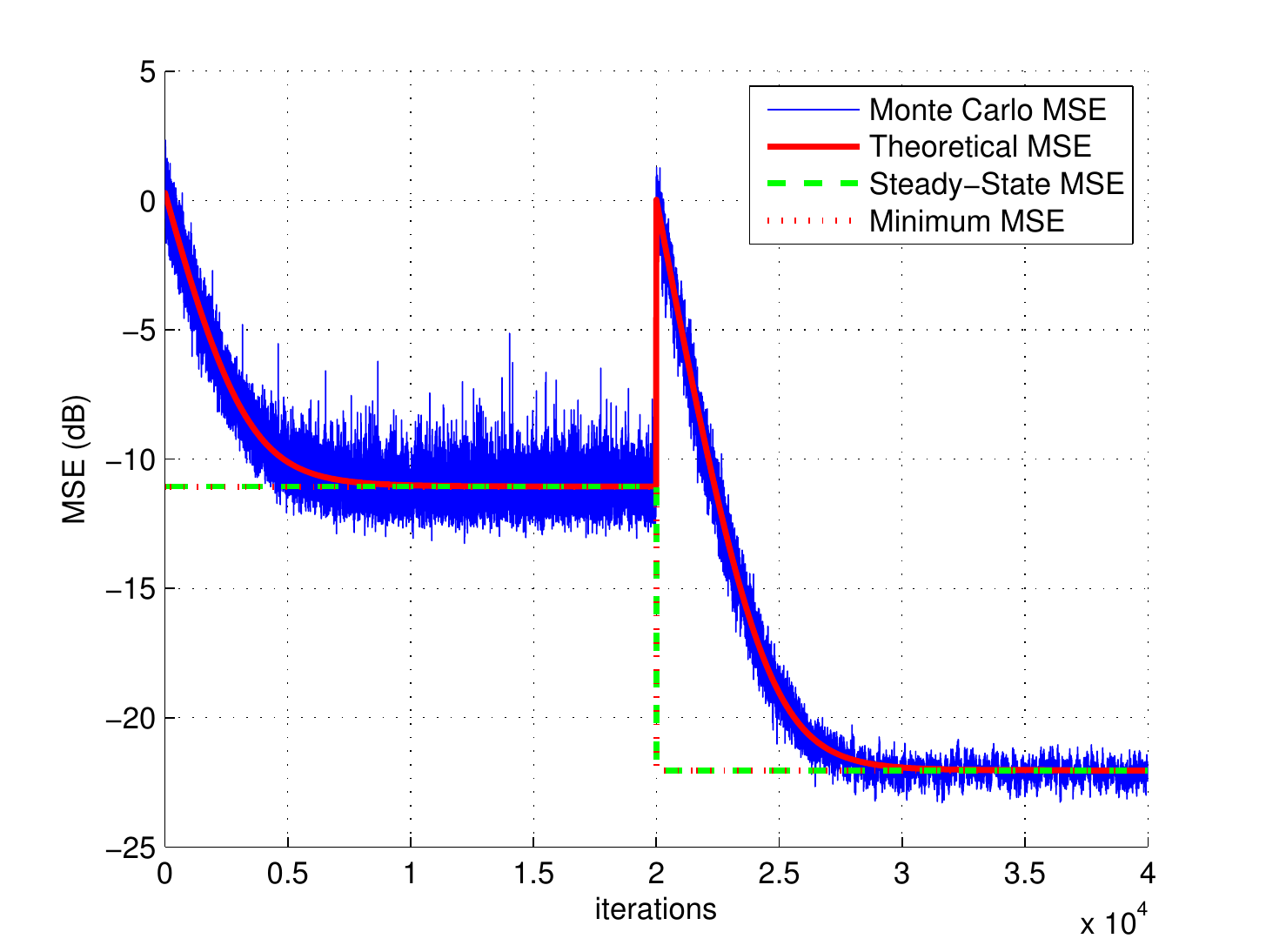}
	\includegraphics[width=5.5cm]{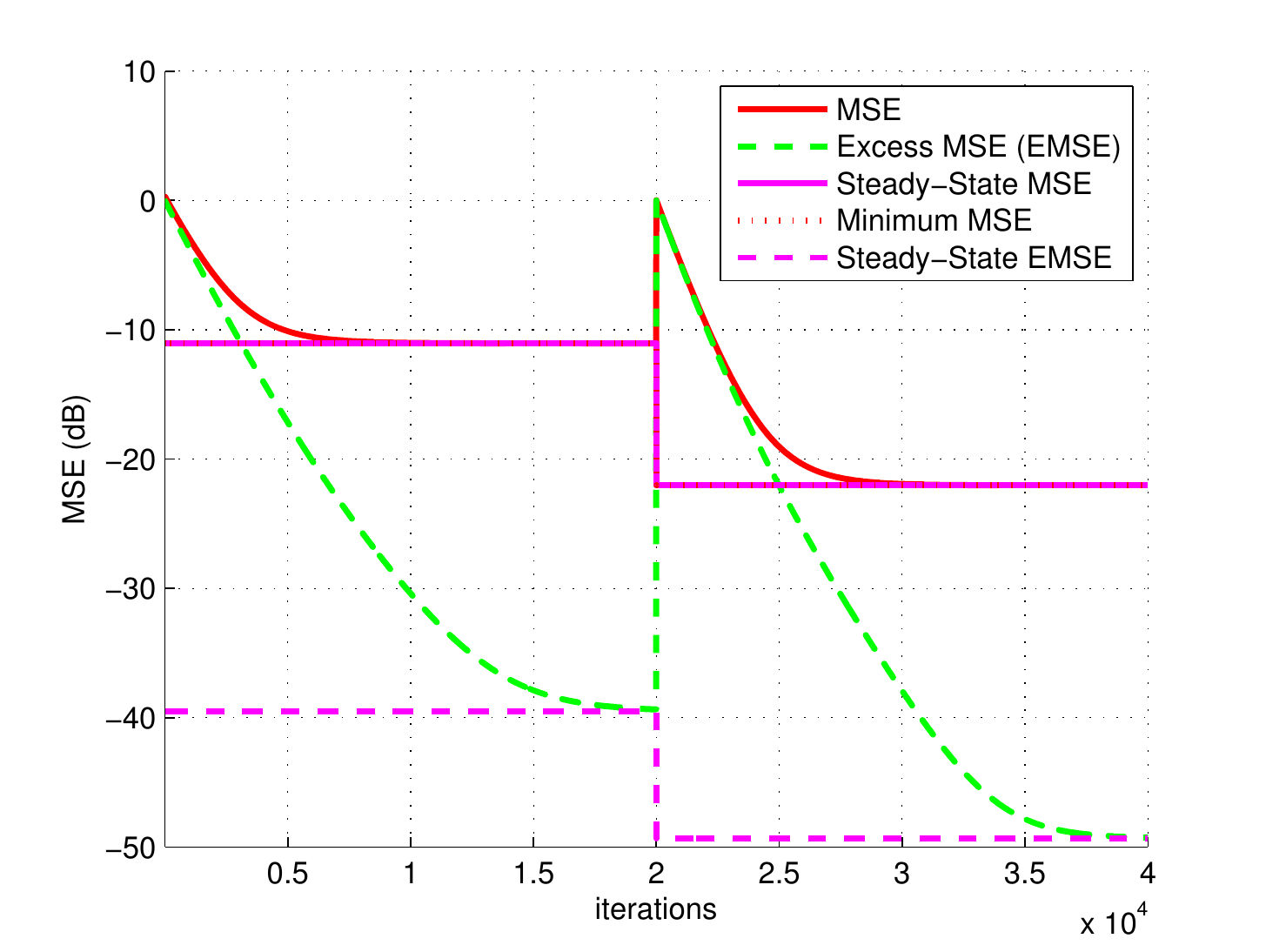}
\end{minipage}}
\subfigure[$\cp{D}_2=\{10@0.15\}$]
{\begin{minipage}[b]{0.333\textwidth}
	\includegraphics[width=5.5cm]{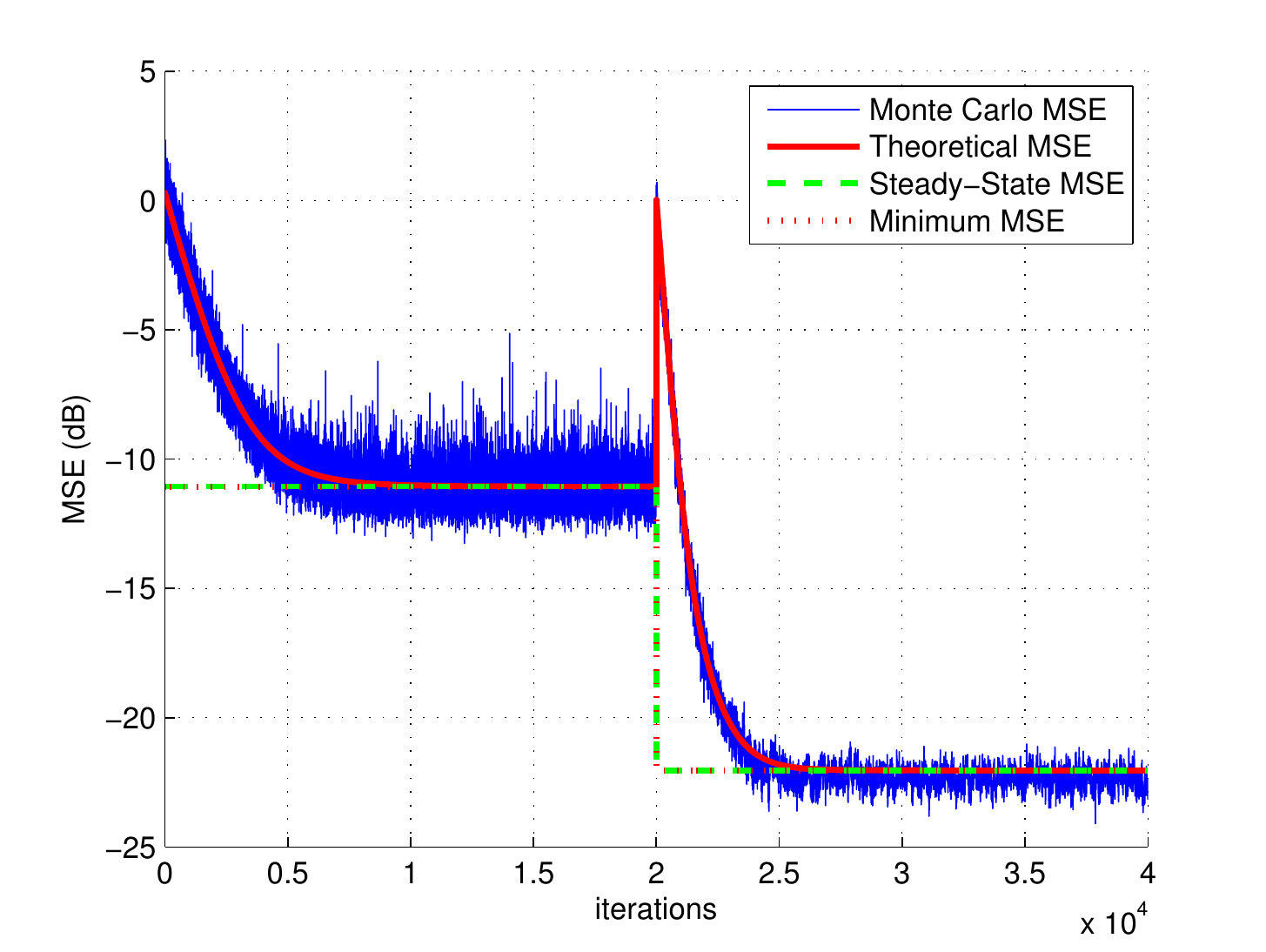}
	\includegraphics[width=5.5cm]{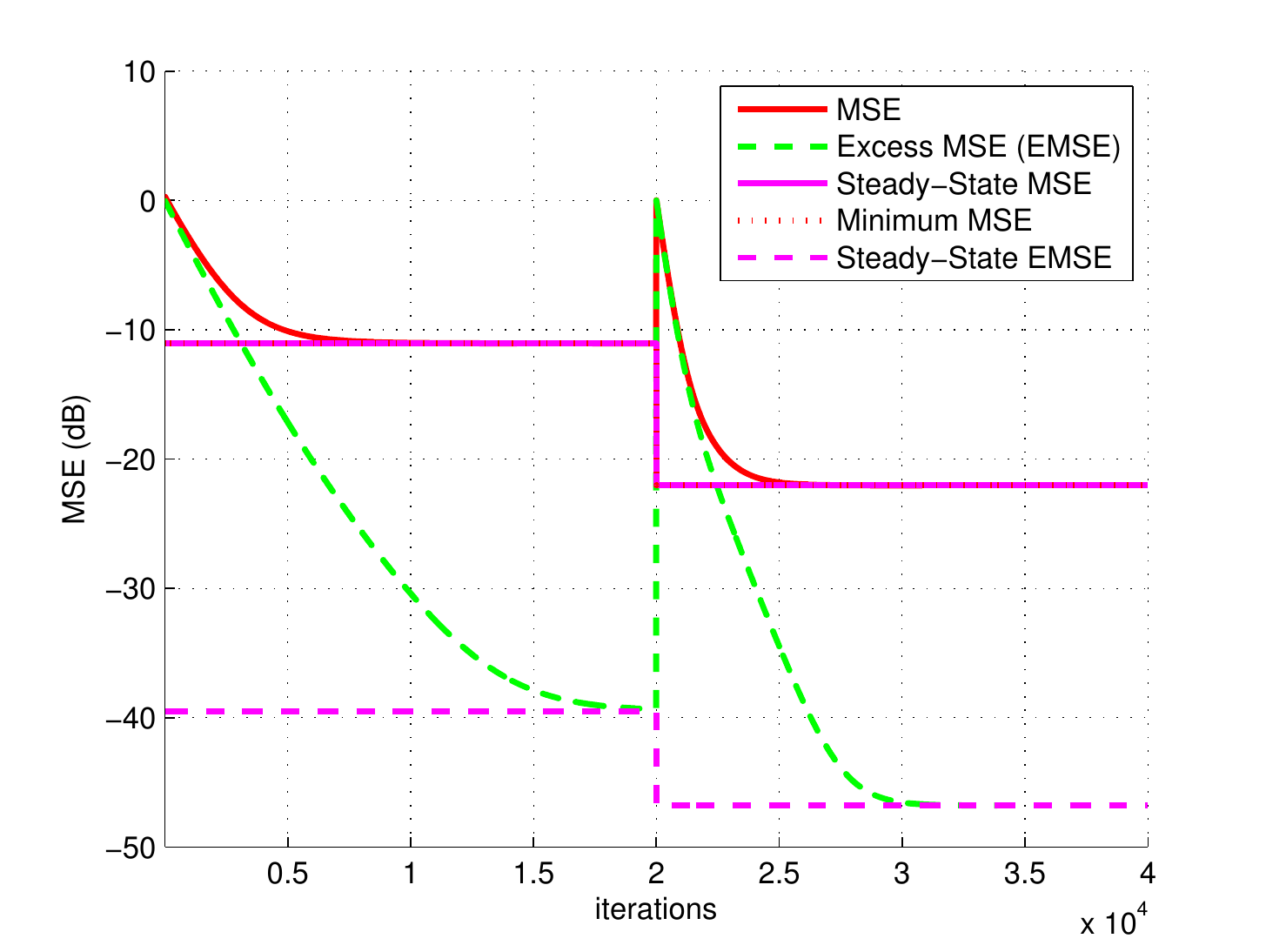}
\end{minipage}}
\subfigure[$\cp{D}_2=\{10@0.15\}\cup\{10@0.35\}$]
{\begin{minipage}[b]{0.333\textwidth}
	\includegraphics[width=5.5cm]{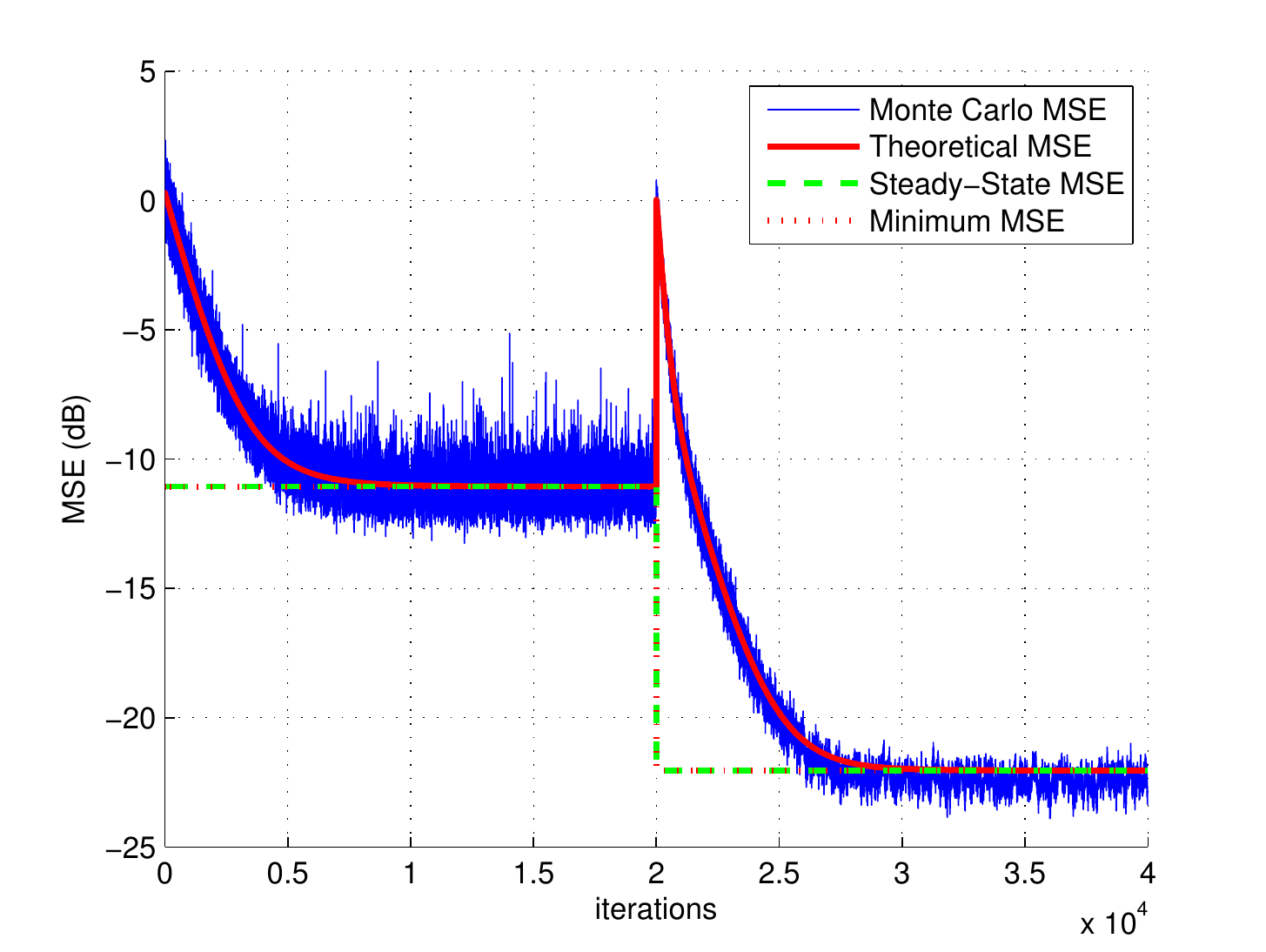}
	\includegraphics[width=5.5cm]{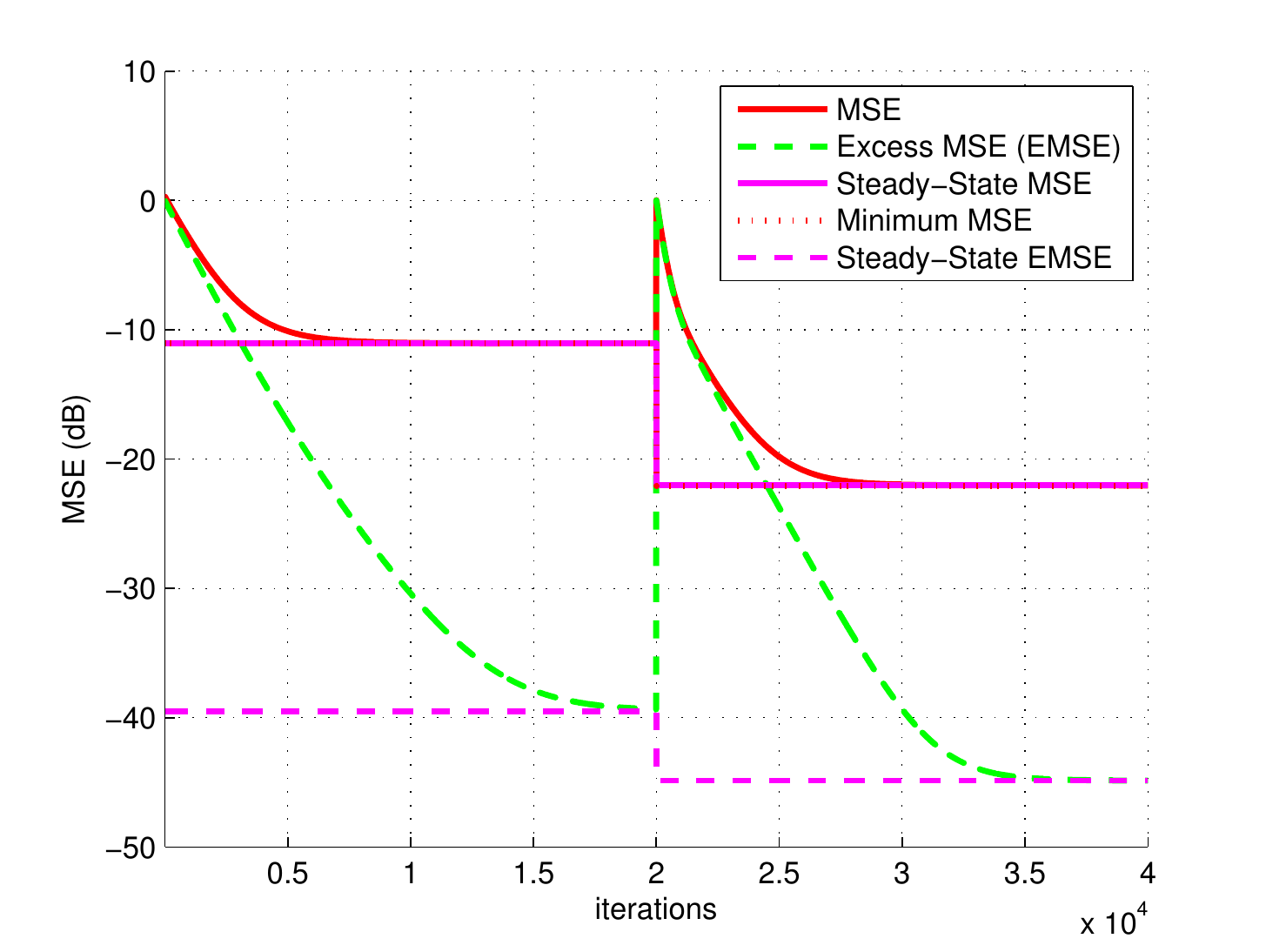}
\end{minipage}}
\caption{Learning curves for Example 1 where $\sigma_u:\,0.35 \rightarrow 0.15$ and $\cp{D}_1=\{10@0.35\}$. See the first row of Table \ref{tab:result_Exp1}.}
\label{fig:Exp1_1}
\end{figure*}

\begin{figure*}[!htb]
\subfigure[$\cp{D}_2=\{10@0.15\}$]
{\begin{minipage}[b]{0.333\textwidth}
	\includegraphics[width=5.5cm]{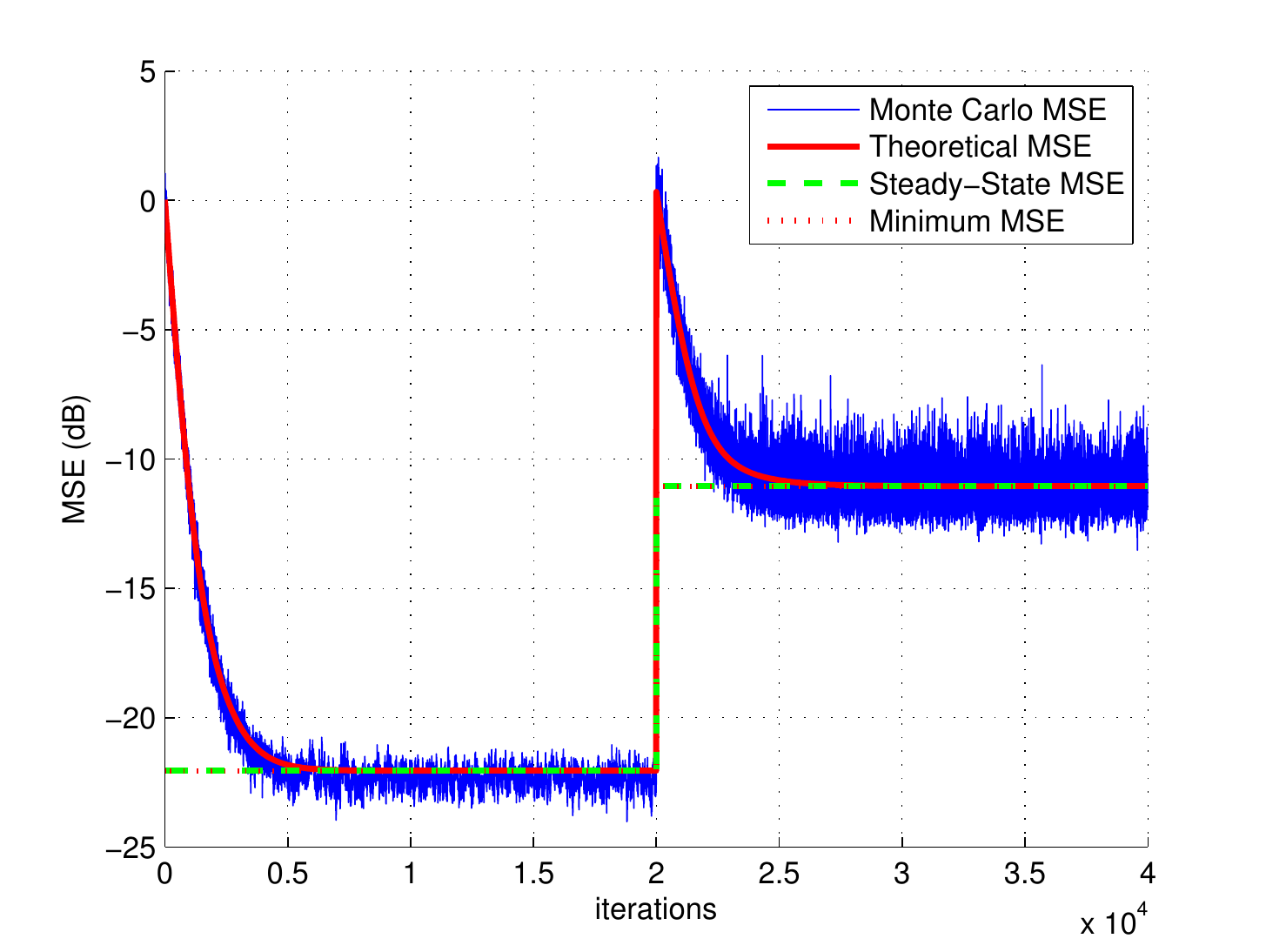}
	\includegraphics[width=5.5cm]{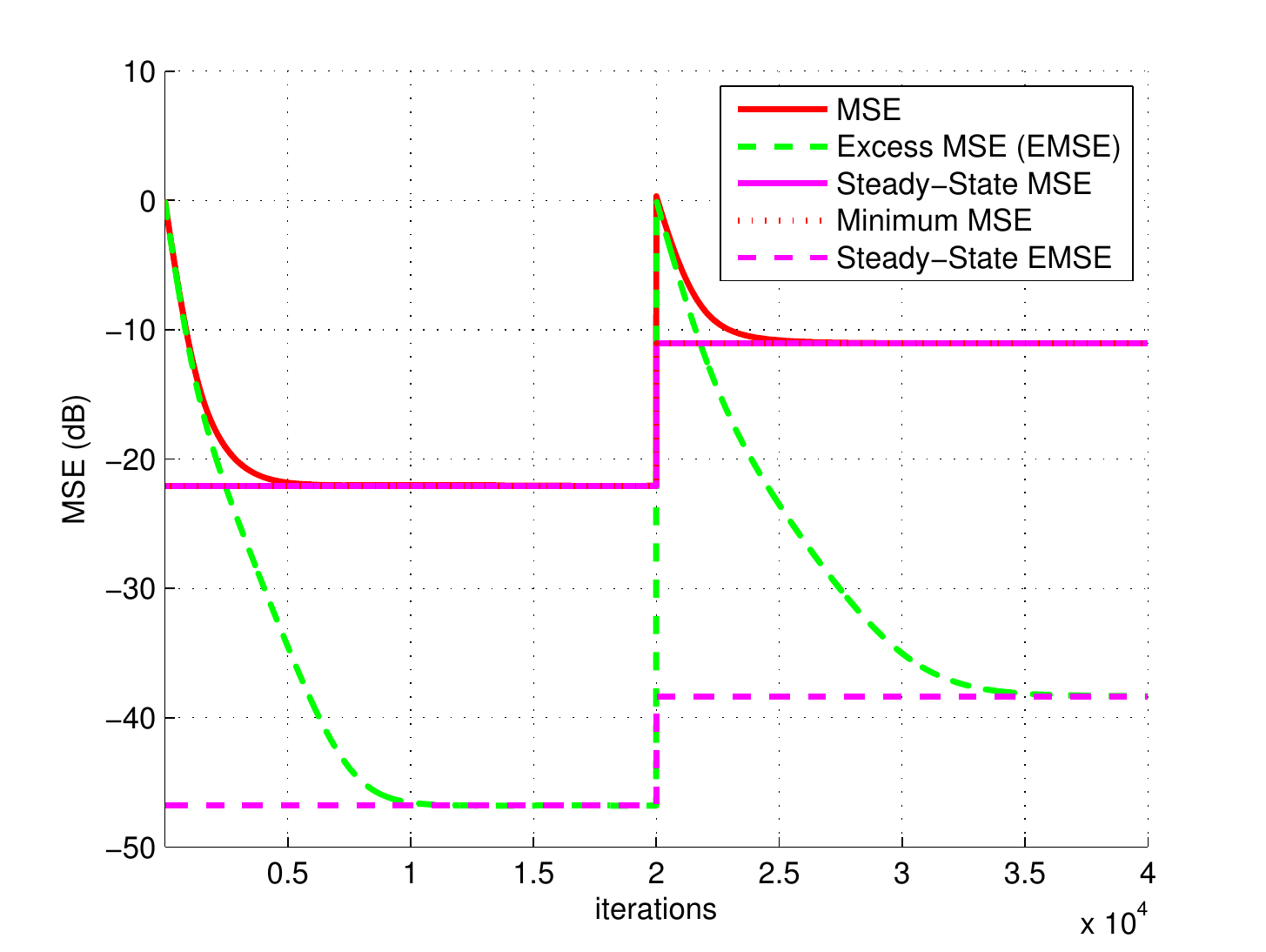}
\end{minipage}}
\subfigure[$\cp{D}_2=\{10@0.35\}$]
{\begin{minipage}[b]{0.333\textwidth}
	\includegraphics[width=5.5cm]{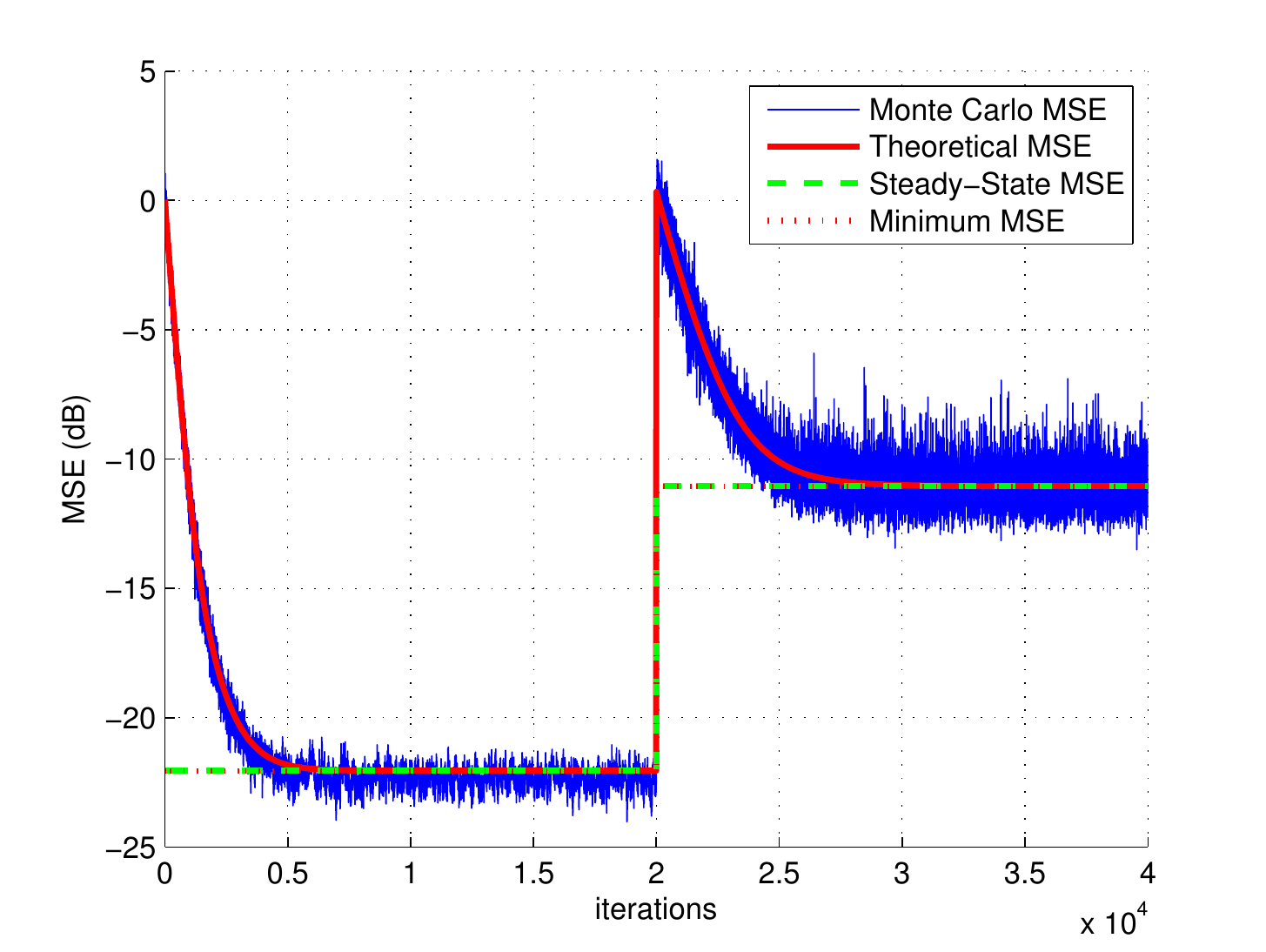}
	\includegraphics[width=5.5cm]{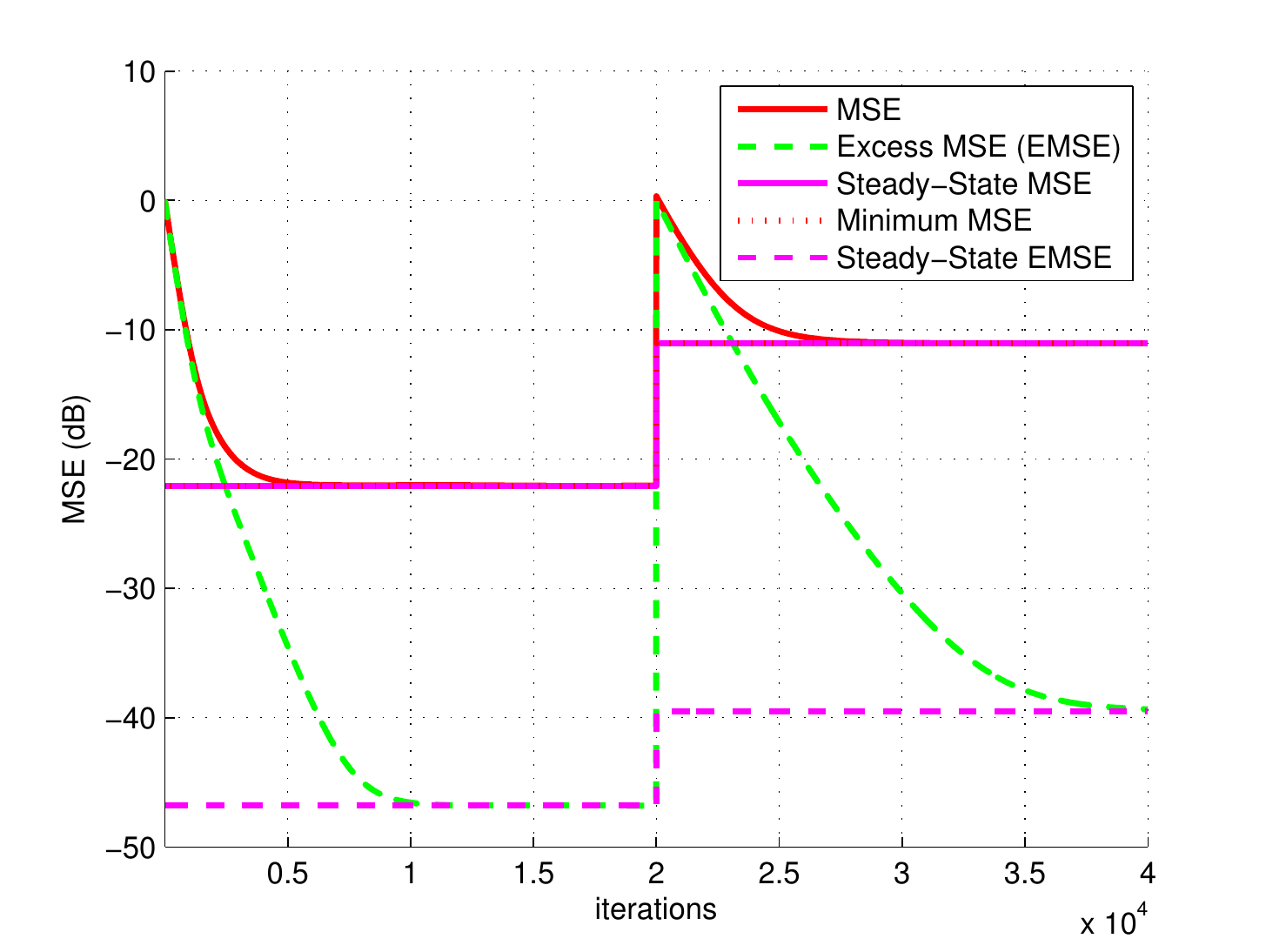}
\end{minipage}}
\subfigure[$\cp{D}_2=\{10@0.15\}\cup\{10@0.35\}$]
{\begin{minipage}[b]{0.333\textwidth}
	\includegraphics[width=5.5cm]{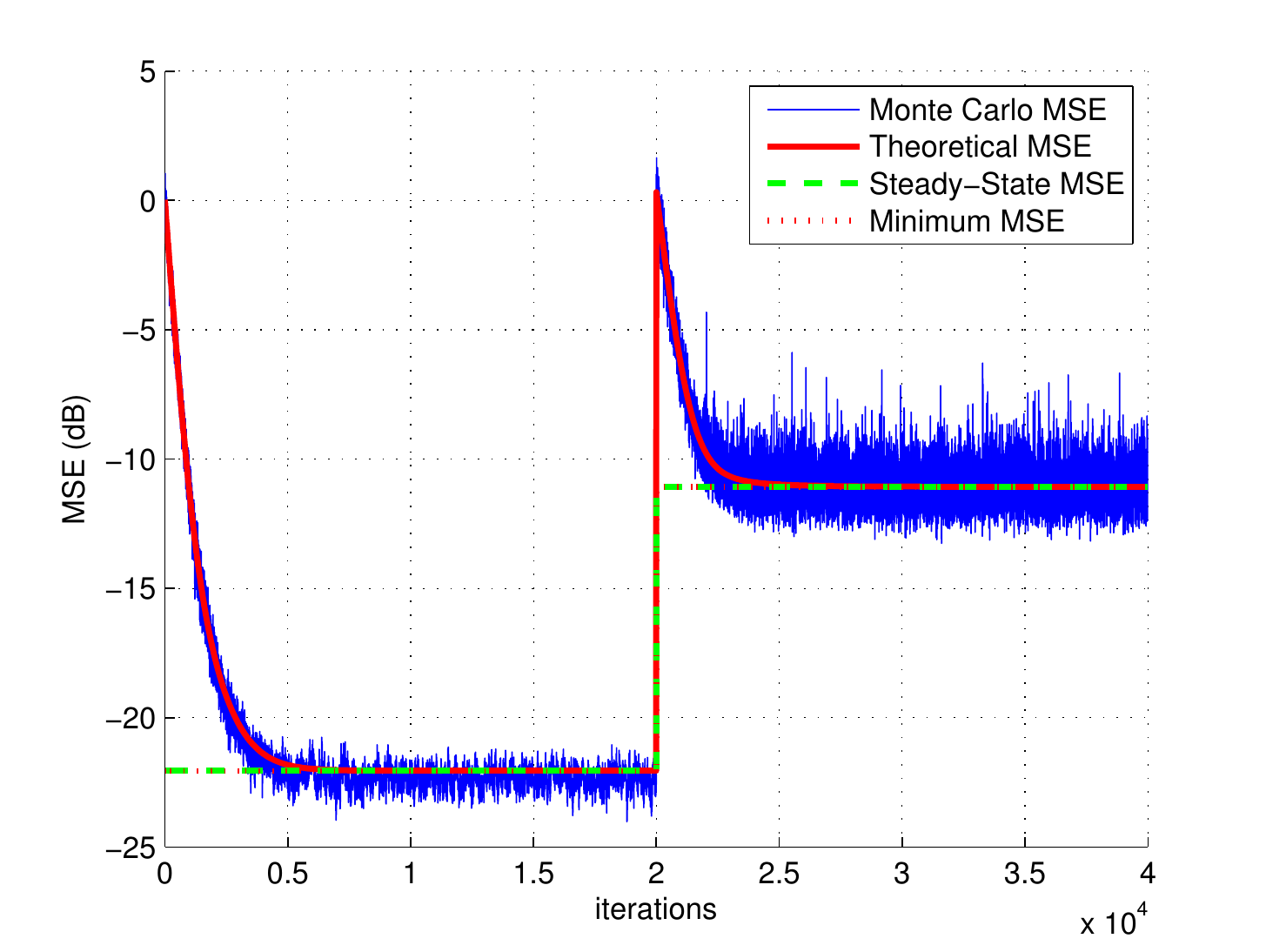}
	\includegraphics[width=5.5cm]{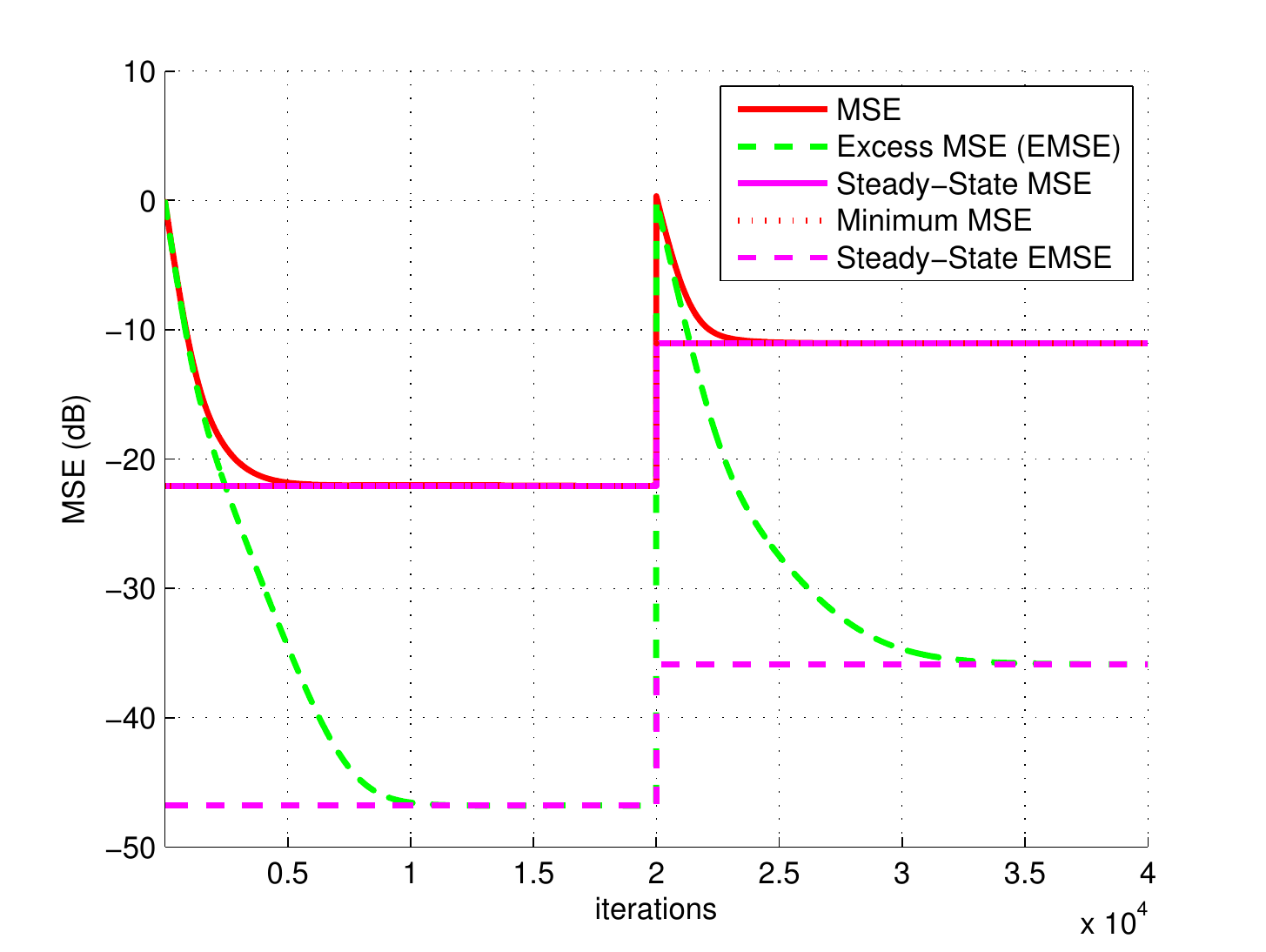}
\end{minipage}}
\caption{Learning curves for Example 1 where $\sigma_u:\,015 \rightarrow 0.35$ and $\cp{D}_1=\{10@0.15\}$. See the second row of Table \ref{tab:result_Exp1}.}
\label{fig:Exp1_2}
\end{figure*}

Table \ref{tab:result_Exp1} presents the simulation conditions, and the experimental results based on $200$ Monte Carlo runs. The convergence iteration number $n_{\epsilon}$ was determined in order to satisfy\begin{equation}
	\label{eq:n_epsilon}
	\|\vc(\infty) - \vc(n_{\epsilon})\|  \leq 10^{-3}.
\end{equation}
Note that $\Jmin$, $\Jmse(\infty)$, $\Jex(\infty)$ and $n_{\epsilon}$ concern convergence in the second subsequence, with the dictionary $\cp{D}_2$. The learning curves are depicted in Figures \ref{fig:Exp1_1} and \ref{fig:Exp1_2}.

\begin{table*}[!htb]
\centering
\caption{Summary of simulation results for Example~1.}
\begin{tabular}{|c|c|c|c|c|c|c|c|c|}\hline
$\xi$  & $\eta$  & $\sigma_u$ & $\cp{D}_1$ & $\cp{D}_2$ & $\Jmin$& $ \Jmse(\infty)$& $\Jex(\infty)$& $n_{\epsilon}$ \\ 
&&&&&[dB]&[dB]& [dB] &  \\ \hline\hline
	&&&&$\{10@0.35\}$&-22.04&-22.03&-49.33&32032 \\ \cline{5-9}
	0.02&0.01&$0.35 \rightarrow 0.15$&$\{10@0.35\}$&$\{10@0.15\}$&\bf{-22.50}&\bf{-22.49}&-47.25&26538 \\ \cline{5-9}
	&&&&$\{10@0.15\}\cup\{10@0.35\}$&-21.90&-21.87&-44.71&30889 \\ \hline \hline
	&&&&$\{10@0.15\}$&-10.98&-10.97&-38.26&32509 \\ \cline{5-9}
	0.02&0.01&$0.15 \rightarrow 0.35$&$\{10@0.15\}$&$\{10@0.35\}$&\bf{-11.20}&\bf{-11.19}&-39.64&36061 \\ \cline{5-9}
	&&&&$\{10@0.15\}\cup\{10@0.35\}$&-11.01&-10.99&-35.81&31614 \\ \hline
\end{tabular}
\label{tab:result_Exp1}
\end{table*}

\subsubsection{Example 2}

Consider the nonlinear dynamic system studied in \cite{Parreira2012,Voros2003} where the input signal was a sequence of statistically independent vectors
\begin{equation}
\label{eq:def_un}
	\vun=[u_1(n)\; u_2(n)]^{\top}
\end{equation}
with correlated samples satisfying $u_1(n)=0.5 u_2(n) + v_u(n)$. The second component of $\vun$, and $v_u(n)$, were i.i.d. zero-mean Gaussian sequences with standard deviation both equal to $\sqrt{0.0656}$, or to $\sqrt{0.0156}$, during the two subsequences of input data. We considered the linear system with memory defined by
\begin{equation}
	y(n)= {\bs a}^\top\, \vun - 0.2\, y(n-1) + 0.35\, y(n-2)
\end{equation}
where ${\bs a } = [1\; 0.5]^{\top}$ and a nonlinear Wiener function
\begin{align}
 \hspace{-0.6mm}       \varphi(y(n))\! &= \! \begin{cases}
                     \displaystyle\frac{y(n)}{3 [0.1\, +\, 0.9\, y^2(n)]^{1/2}} \!&\text{for}\ y(n)\! \geq\! 0 \\
                     \displaystyle\frac{-y^2(n)[1- \exp(0.7 y(n))]}{3}        \! &\text{for}\ y(n) \!< \!0
\end{cases} \\
   d(n) &= \varphi(y(n)) + z(n)
\end{align}
where $d(n)$ is the output signal. It was corrupted by a zero-mean i.i.d. Gaussian noise $z(n)$ with variance $\sigma_z^2 = 10^{-6}$. The initial condition {$y(1) = 0$} was considered. The bandwidth $\xi$ of the Gaussian kernel was set to $0.05$, and the step-size $\eta$ of the KLMS was set to $0.05$. The length of each input sequence was $4\times 10^4$. As in Example 1, two changes were considered. For the first one, the standard deviation of $u_2(n)$ and $v_u(n)$ was changed from $\sqrt{0.0656}$  to $\sqrt{0.0156}$ at time instant $n=1\times 10^4$. Conversely, for the second one, it was changed from $\sqrt{0.0156}$  to $\sqrt{0.0656}$.

\begin{figure*}[!htb]
\subfigure[$\cp{D}_2=\{15@\sqrt{0.0656}\}$]
{\begin{minipage}[b]{0.333\textwidth}
	\includegraphics[width=5.5cm]{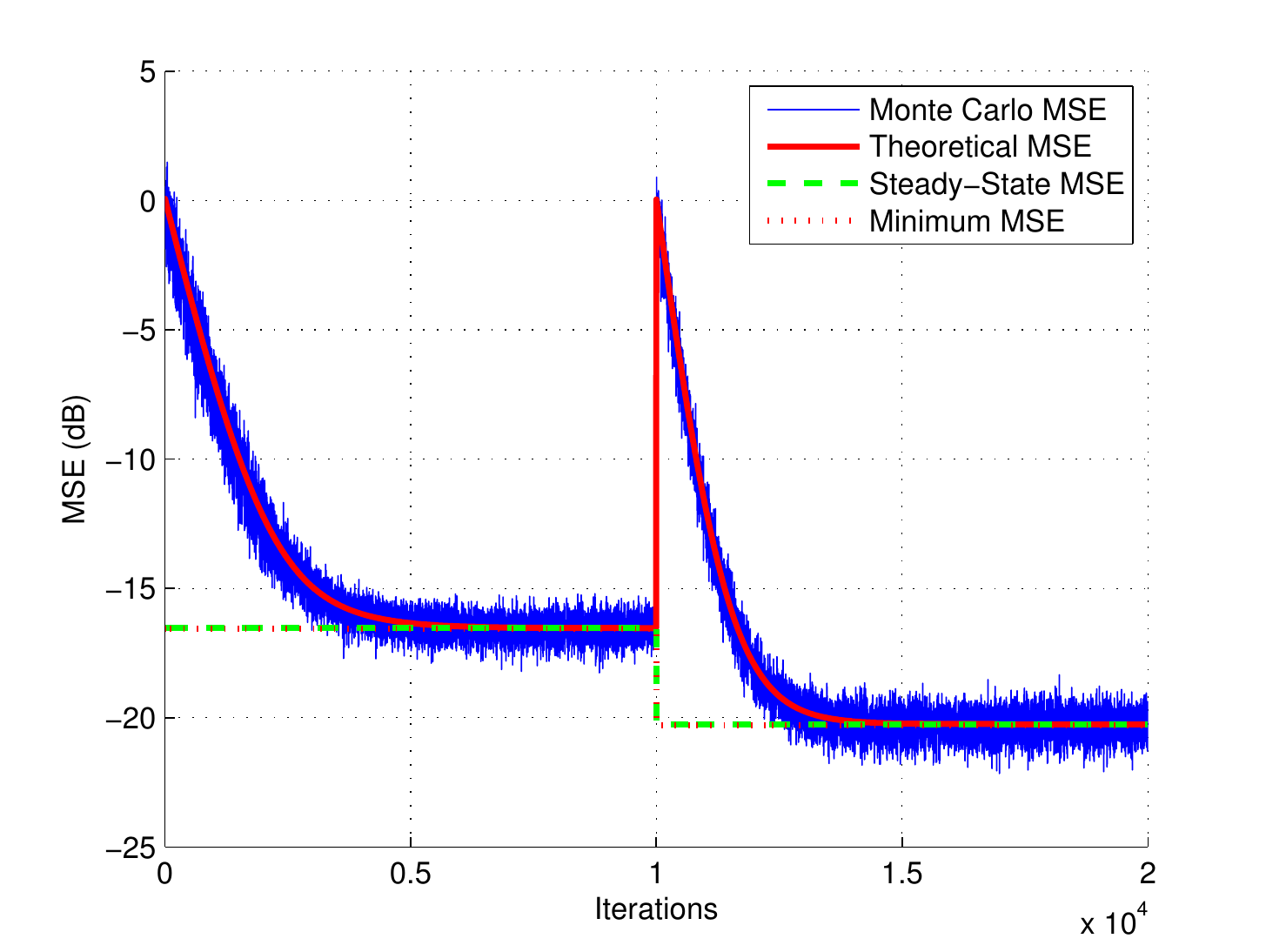}
	\includegraphics[width=5.5cm]{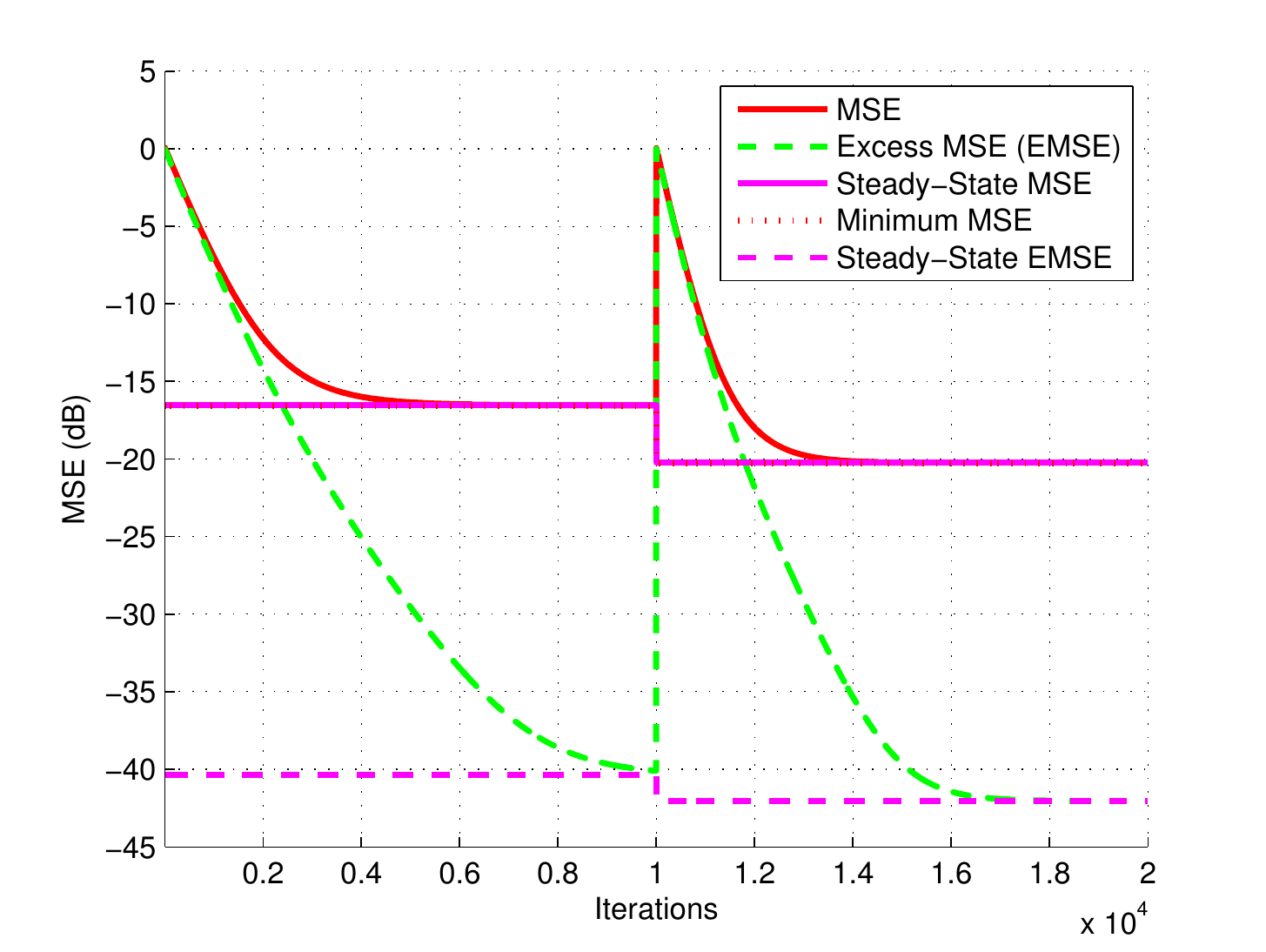}
\end{minipage}}
\subfigure[$\cp{D}_2=\{15@\sqrt{0.0156}\}$]
{\begin{minipage}[b]{0.333\textwidth}
	\includegraphics[width=5.5cm]{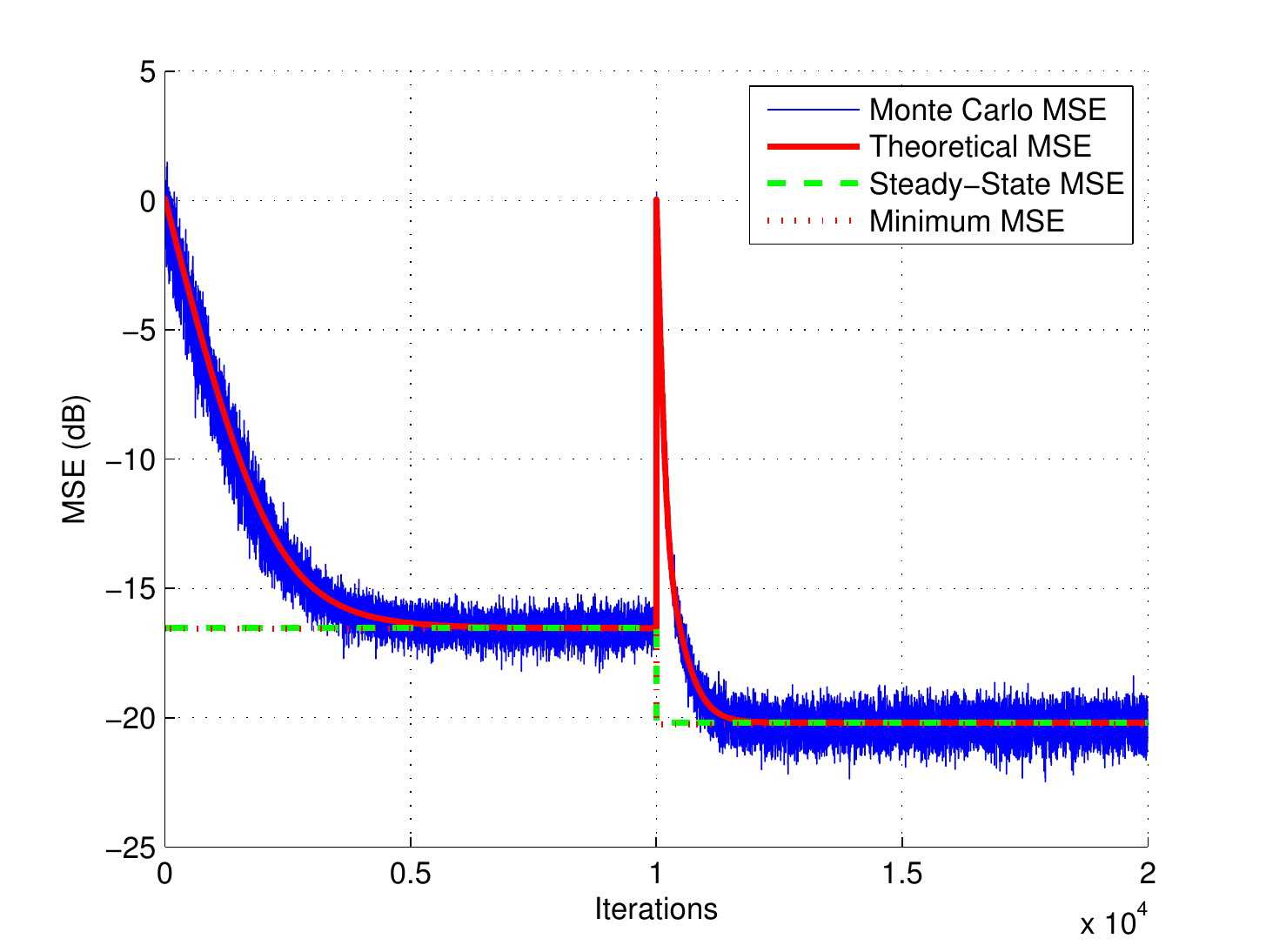}
	\includegraphics[width=5.5cm]{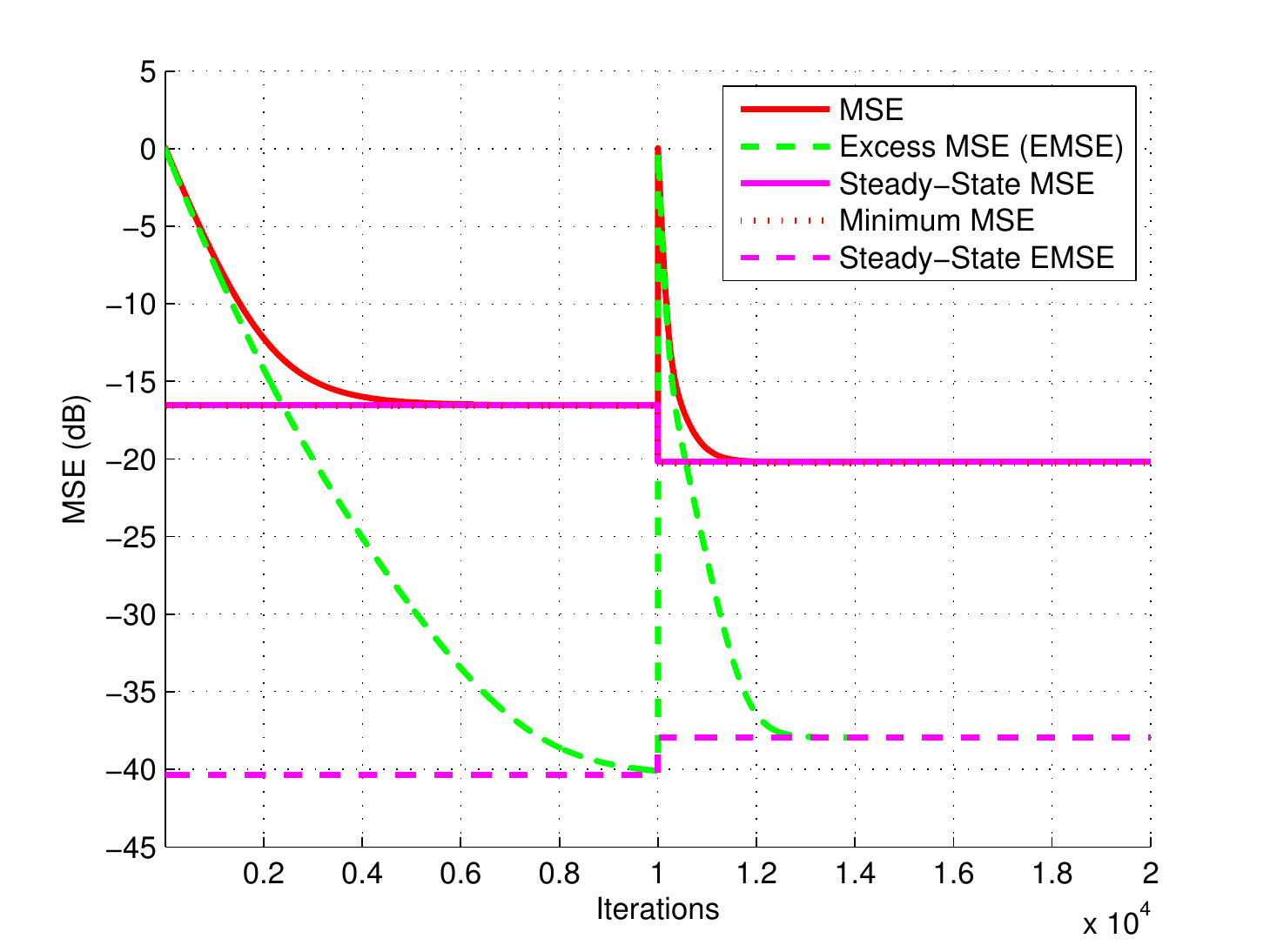}
\end{minipage}}
\subfigure[$\cp{D}_2=\{15@\sqrt{0.0156}\}\cup\{15@\sqrt{0.0656}\}$]
{\begin{minipage}[b]{0.333\textwidth}
	\includegraphics[width=5.5cm]{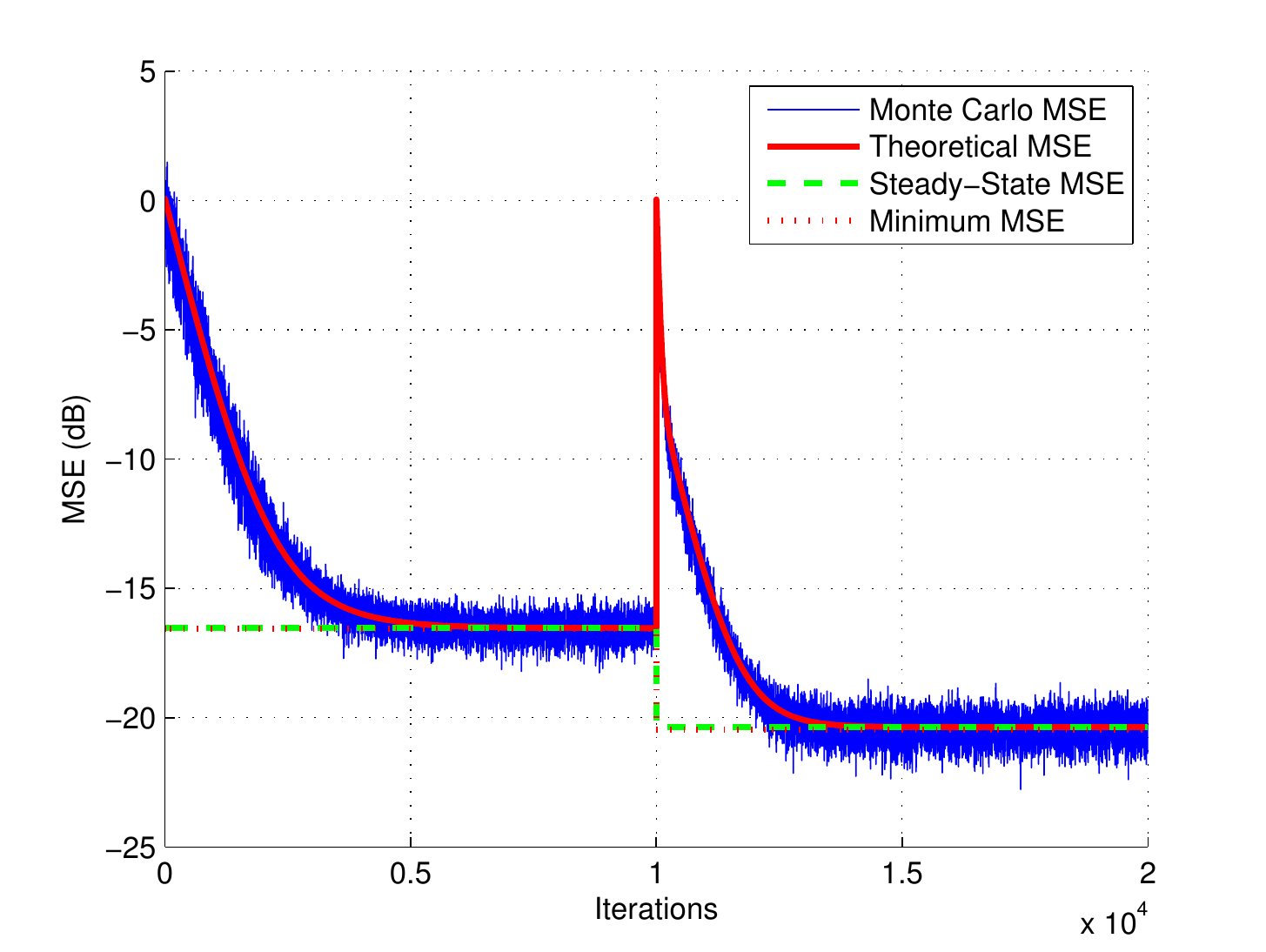}
	\includegraphics[width=5.5cm]{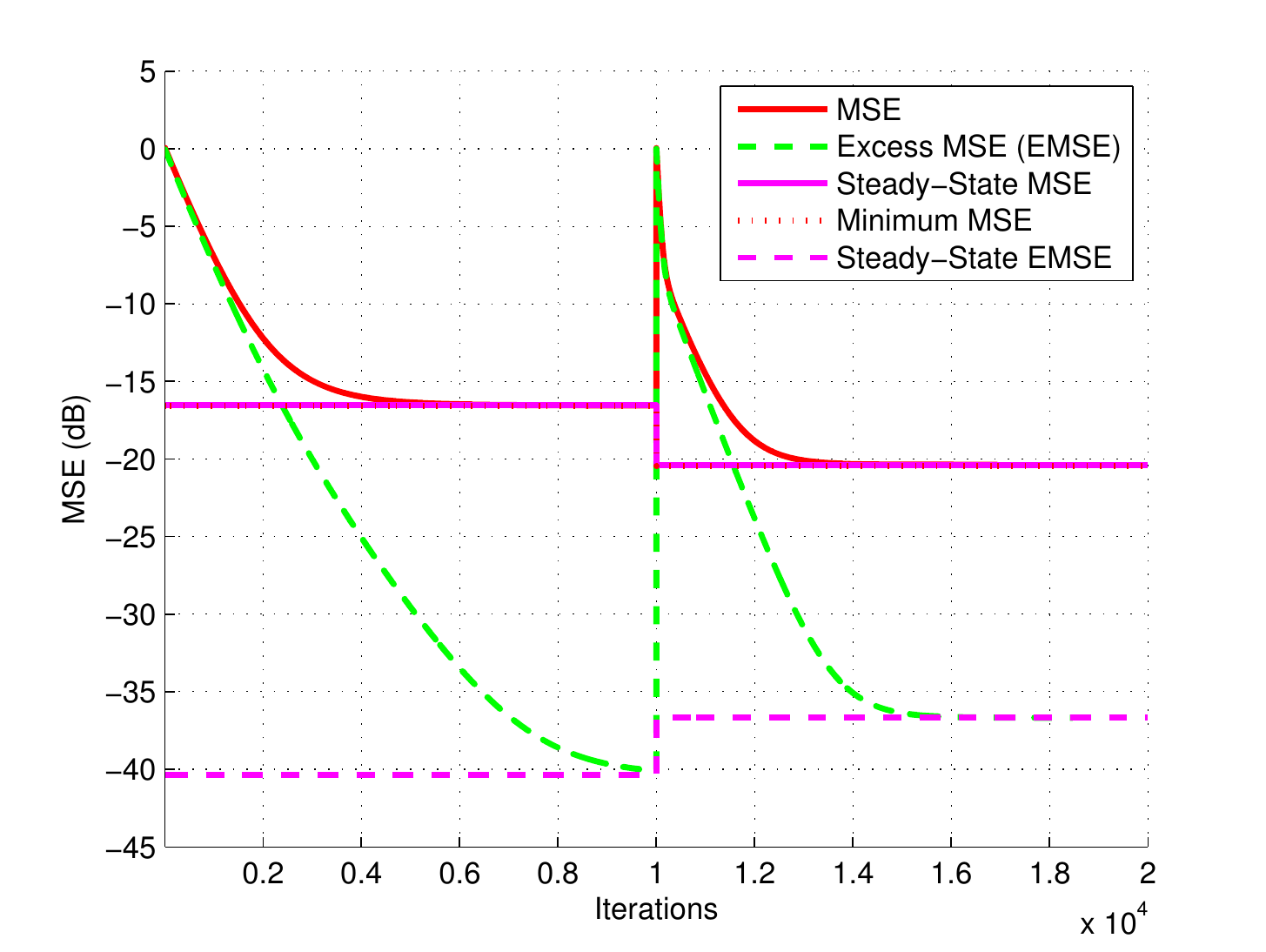}
\end{minipage}}
\caption{Learning curves for Example 2 with $\sigma_{u_2},\sigma_{v_u}:\,\sqrt{0.0656} \rightarrow \sqrt{0.0156}$ and $\cp{D}_1=\{15@\sqrt{0.0656}\}$. See the first row of Table \ref{tab:result_Exp2}.}
\label{fig:Exp2_1}
\end{figure*}

\begin{figure*}[!htb]
\subfigure[$\cp{D}_2=\{15@\sqrt{0.0156}\}$]
{\begin{minipage}[b]{0.333\textwidth}
	\includegraphics[width=5.5cm]{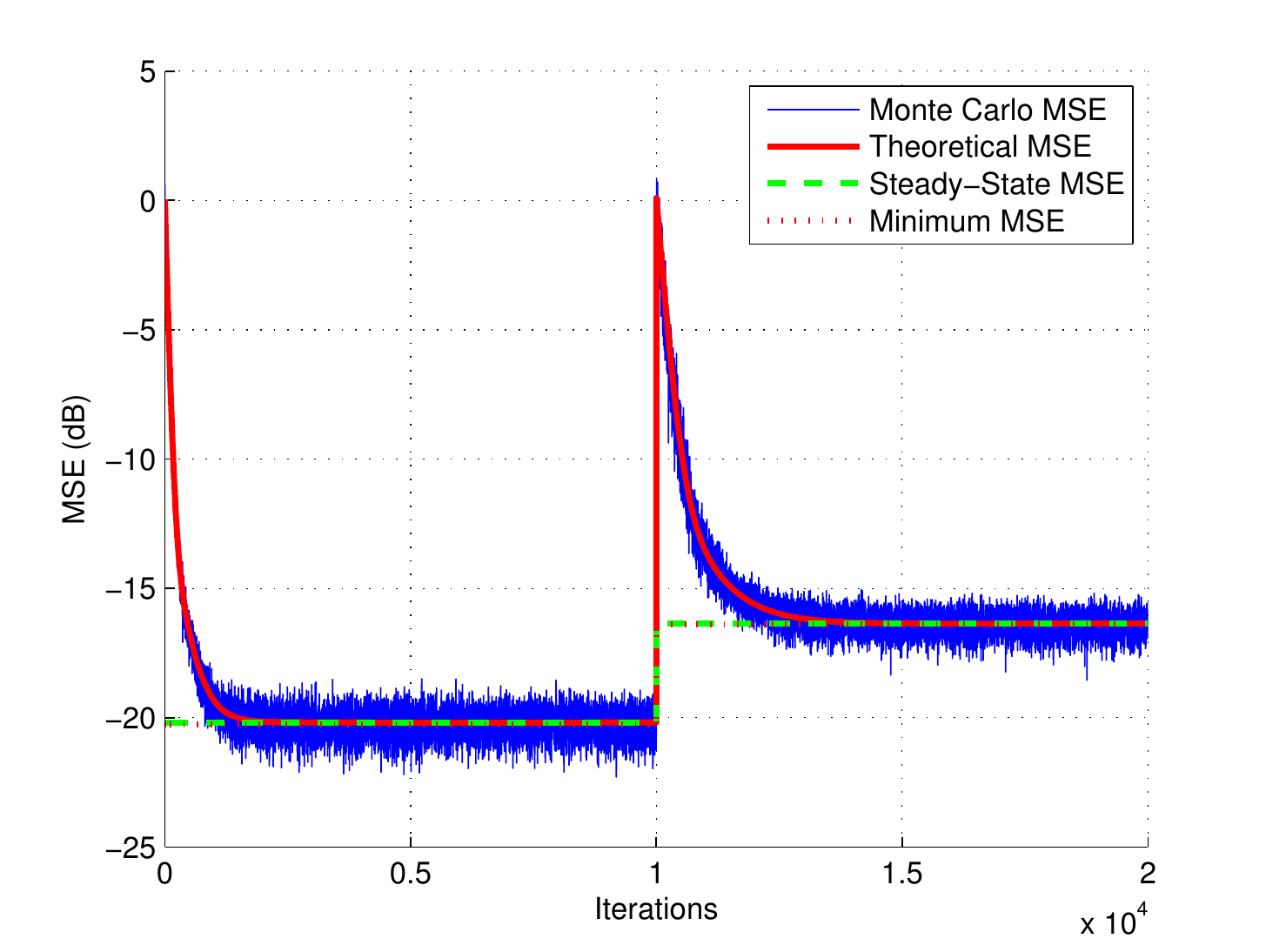}
	\includegraphics[width=5.5cm]{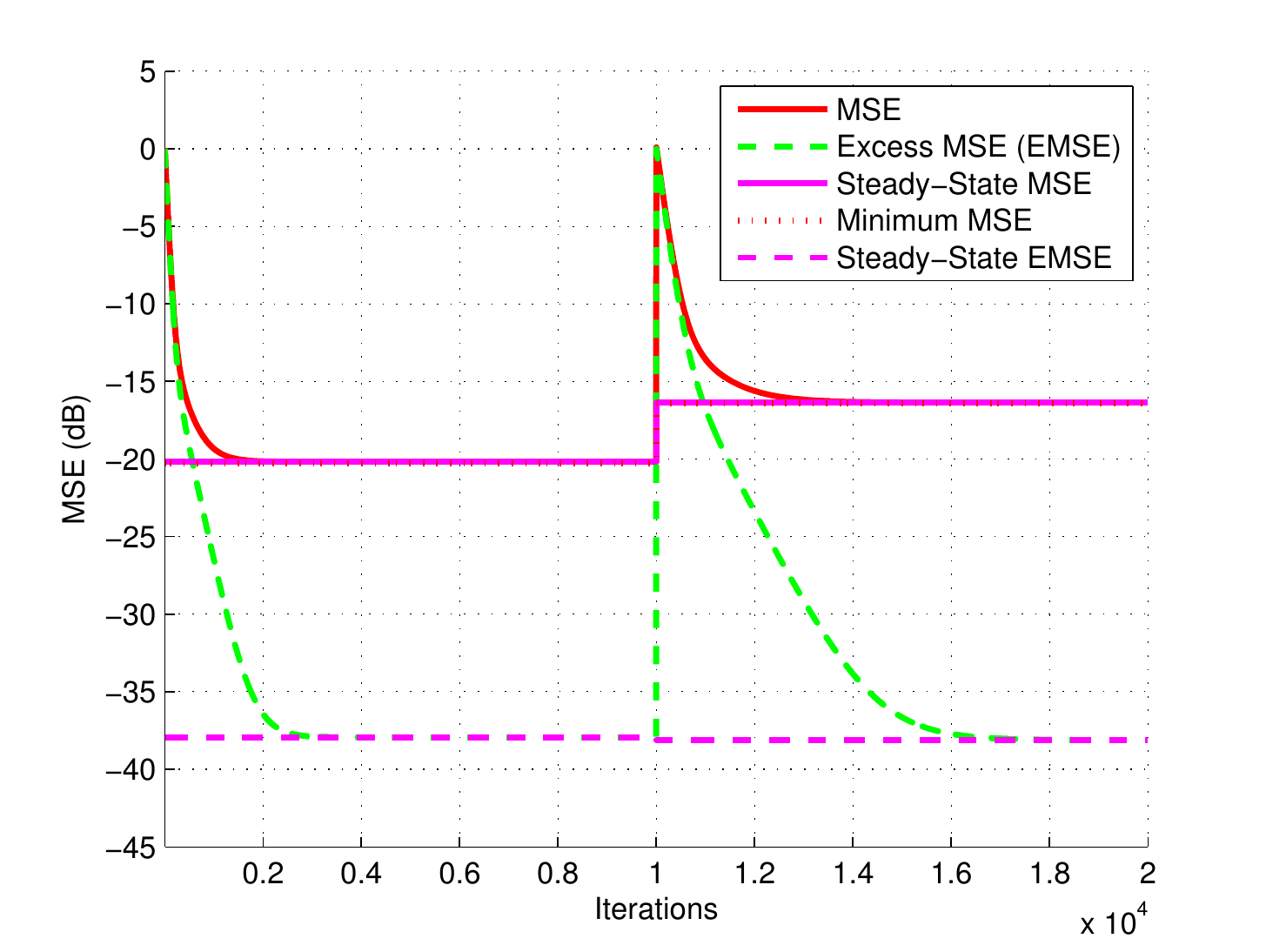}
\end{minipage}}
\subfigure[$\cp{D}_2=\{15@\sqrt{0.0656}\}$]
{\begin{minipage}[b]{0.333\textwidth}
	\includegraphics[width=5.5cm]{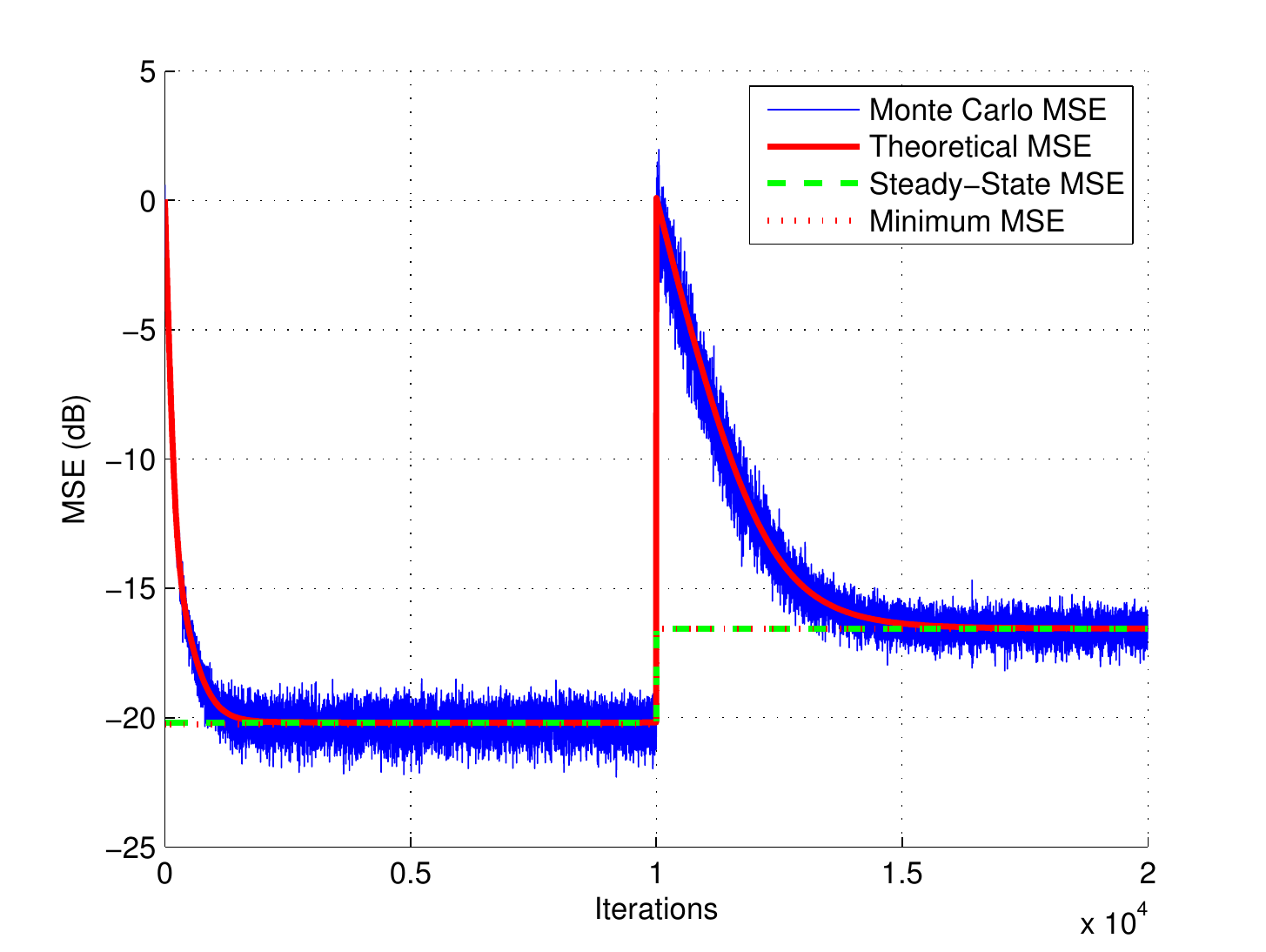}
	\includegraphics[width=5.5cm]{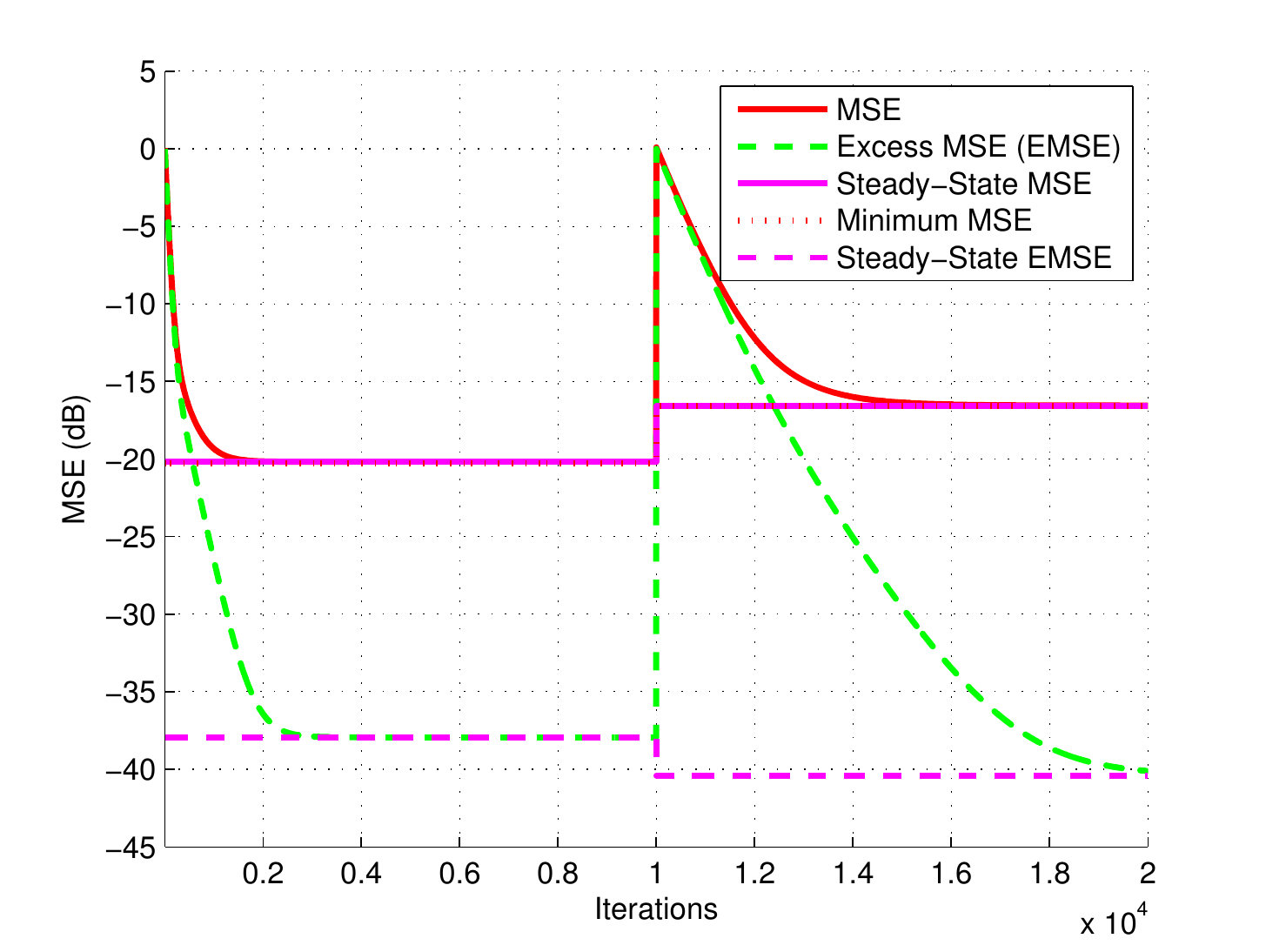}
\end{minipage}}
\subfigure[$\cp{D}_2=\{15@\sqrt{0.0156}\}\cup\{15@\sqrt{0.0656}\}$]
{\begin{minipage}[b]{0.333\textwidth}
	\includegraphics[width=5.5cm]{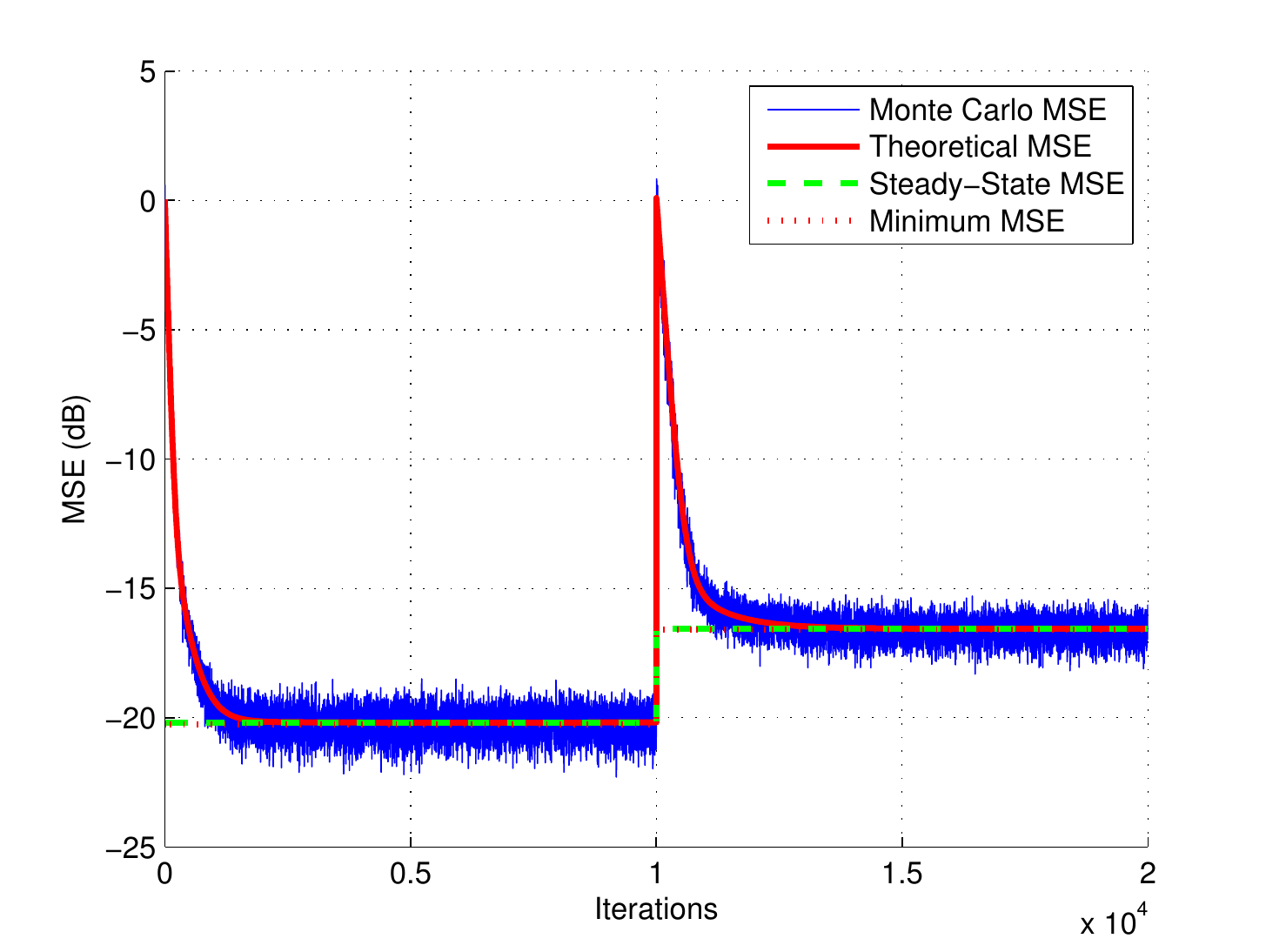}
	\includegraphics[width=5.5cm]{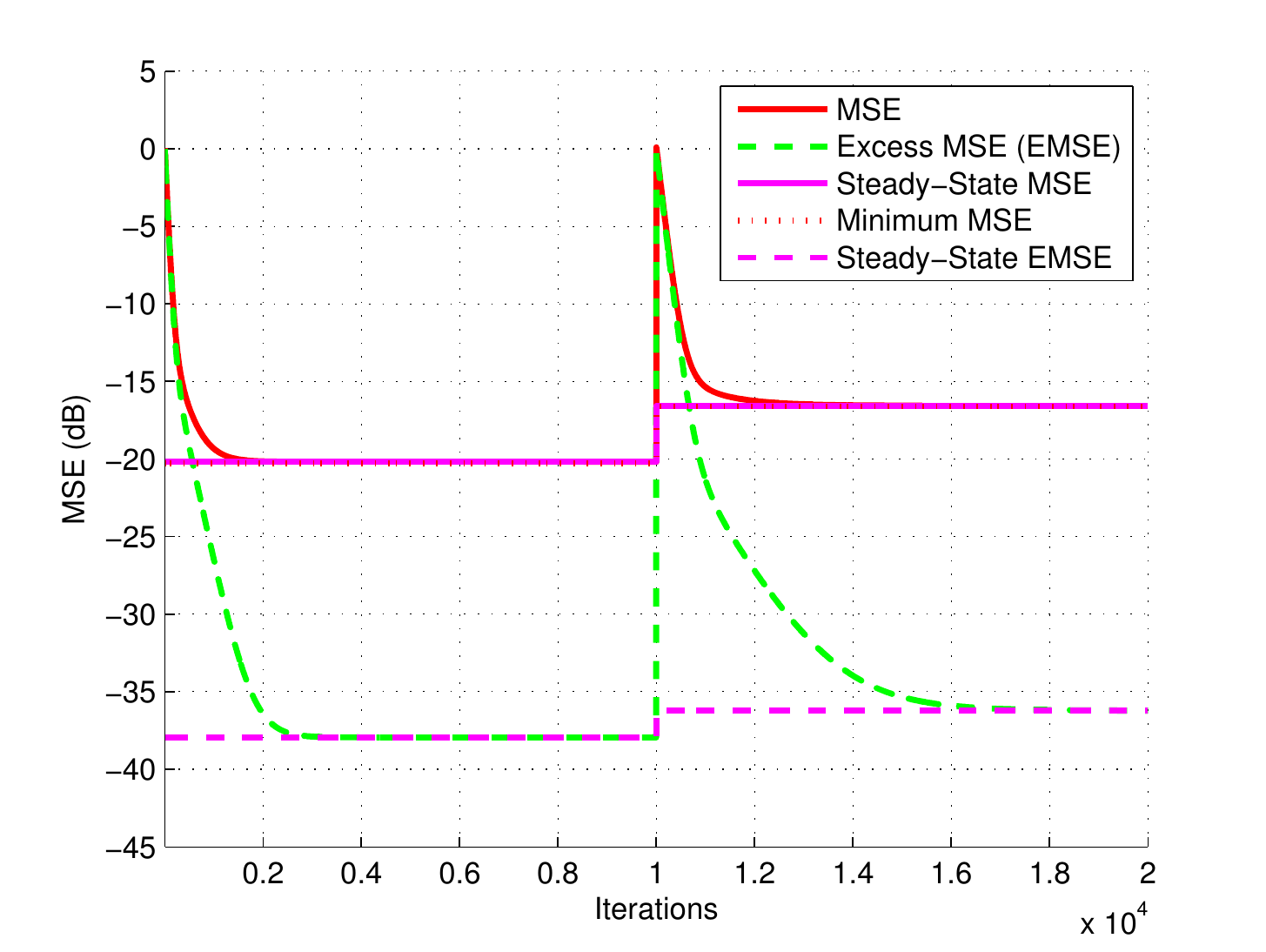}
\end{minipage}}
\caption{Learning curves for Example 2 with $\sigma_{u_2},\sigma_{v_u}:\,\sqrt{0.0156} \rightarrow \sqrt{0.0656}$ and $\cp{D}_1=\{15@\sqrt{0.0156}\}$. See the second row of Table \ref{tab:result_Exp2}.}
\label{fig:Exp2_2}
\end{figure*}

Table \ref{tab:result_Exp2} presents the results based on $200$ Monte Carlo runs. Note that $\Jmin$, $\Jmse(\infty)$, $\Jex(\infty)$ and $n_{\epsilon}$ concern convergence in the second subsequence, with dictionary $\cp{D}_2$. The learning curves are depicted in Figures \ref{fig:Exp2_1} and \ref{fig:Exp2_2}.

\begin{table*}[!htb]
\centering
\caption{{Summary of simulation results for Example 2}.}
\begin{tabular}{|c|c|c|c|c|c|c|c|c|}\hline
$\xi$  & $\eta$  & $\sigma_{u_2},\sigma_{v_u}$ & $\cp{D}_1$ & $\cp{D}_2$ & $\Jmin$& $ \Jmse(\infty)$& $\Jex(\infty)$& $n_{\epsilon}$ \\ 
&&&&&[dB]&[dB]& [dB] &  \\ \hline\hline
	&&&&$\{15@\sqrt{0.0656}\}$&-20.28&-20.25&-42.04&15519 \\ \cline{5-9}
	0.05&0.05&$\sqrt{0.0656} \rightarrow \sqrt{0.0156}$&$\{15@\sqrt{0.0656}\}$&$\{15@\sqrt{0.0156}\}$&-20.27&-20.20&-37.96&12117 \\ \cline{5-9}
	&&&&$\{15@\sqrt{0.0156}\}\cup\{15@\sqrt{0.0656}\}$&\bf{-20.47}&\bf{-20.37}&-36.68&14731 \\ \hline \hline
	&&&&$\{15@\sqrt{0.0156}\}$&-16.40&-16.37&-38.12&15858 \\ \cline{5-9}
	0.05&0.05&$\sqrt{0.0156} \rightarrow \sqrt{0.0656}$&$\{15@\sqrt{0.0156}\}$&$\{15@\sqrt{0.0656}\}$&-16.57&-16.55&-40.39&19269 \\ \cline{5-9}
	&&&&$\{15@\sqrt{0.0156}\}\cup\{15@\sqrt{0.0656}\}$&\bf{-16.61}&\bf{-16.57}&-36.21&16123 \\ \hline
\end{tabular}
\label{tab:result_Exp2}
\end{table*}

\subsubsection{Discussion}

We shall now discuss the simulation results. It is important to recognize the significance of the mean-square estimation errors provided by the model, which perfectly match the averaged Monte Carlo simulation results. The model separates the contribution of the minimum MSE and EMSE, and makes comparisons possible. The simulation results clearly show that adjusting the dictionary to the input signal has a positive effect on the performance when a change in the statistics is detected. This can be done by adding new elements to the existing dictionary, while at the same time possibly discarding the obsolete elements. Considering a completely new dictionary led us to the lowest MSE $J_\text{ms}(\infty)$ and minimum MSE $J_\text{min}$ in Example 1. Adding new elements to the existing dictionary provided the lowest MSE $J_\text{ms}(\infty)$ and minimum MSE $J_\text{min}$ in Example 2. This strategy can however have a negative effect on the convergence behavior of the algorithm. As a conclusion, the simulation results clearly show the need for an online dictionary update mechanism.

\section{KLMS algorithm with forward-backward splitting}

We shall now introduce a KLMS-type algorithm based on forward-backward splitting, which can automatically update the dictionary in an online way by discarding the obsolete elements and adding appropriate ones.

\subsection{Forward-backward splitting method in  a nutshell}
Consider first the following optimization problem
\begin{equation}
        \label{eq:general.fbs}
        \balpha^*=\mathop{\arg\min}_{\balpha\in\R^N} \left\{Q(\balpha)=J(\balpha)+\lambda\Omega(\balpha)\right\}
\end{equation}
where $J(\cdot)$ is a convex empirical loss function with Lipschitz continuous gradient and Lipschitz constant $1/\eta_0$. Function $\Omega(\cdot)$ is a convex, continuous, but not necessarily differentiable regularizer, and $\lambda$ is a regularization constant. This problem has been extensively studied in the literature, and can be solved with forward-backward splitting \cite{Beck2009}. In a nutshell, this approach consists of minimizing the following quadratic approximation of $Q(\balpha)$ at a given point $\balpha_n$, in an iterative way,
\begin{equation}
  	\label{eq:approche-proximale}
	\begin{split}
  	Q_{\eta}(\balpha,\balpha_n) = &J(\balpha_n) +\nabla J(\balpha_n)^\top(\balpha-\balpha_n)\\
	 		& \;+ \frac{1}{2\eta}\|\balpha-\balpha_n\|_2^2+ \lambda\Omega(\balpha)
	\end{split}
\end{equation}
since $Q(\balpha) \leq Q_{\eta}(\balpha,\balpha_n)$ for any $\eta \leq \eta_0$. Simple algebra shows that the function $Q_{\eta}(\balpha,\balpha_n)$ admits a unique minimizer, denoted by $\balpha_{n+1}$, given by
\begin{equation}
	\label{eq:fobos}
	\balpha_{n+1}=\mathop{\arg\min}_{\balpha\in\R^N}\left\{\lambda\Omega(\balpha)+\frac{1}{2\eta}\|\balpha-\hat{\balpha}_{n}\|_2^2\right\}
\end{equation}
with $\hat\balpha_{n}=\balpha_n-\eta\nabla J(\balpha_n)$. It is interesting to note that $\hat\balpha_n$ can be interpreted as an intermediate gradient descent step on the cost function $J(\cdot)$. Problem \eqref{eq:fobos} is called the proximity operator for the regularizer $\Omega(\cdot)$, and is denoted by $\text{Prox}_{{\lambda\eta}\Omega(\cdot)}(\cdot)$. While this method can be considered as a two-step optimization procedure, it is equivalent to a subgradient descent with the advantage of promoting exact sparsity at each iteration. The convergence of the optimization procedure \eqref{eq:fobos} to a global minimum is ensured if $1/\eta$ is a Lipschitz constant of the gradient $\nabla J(\balpha)$.  In the case $J(\balpha)=\frac{1}{2}\|\bd-\bK\balpha\|_2^2$ considered in~\eqref{eq:problem.parametric}, where $\bK$ is a $(N \times N)$ matrix, a well-established condition ensuring the convergence of $\balpha_{n+1}$ to a minimizer of problem \eqref{eq:general.fbs} is to require that \cite{Beck2009} 
\begin{equation}
	\label{eq:upper.bound}
	0<\eta<2/\lambda_{\max}(\bK^\top\bK)
\end{equation}
where $\lambda_{\max}(\cdot)$ is the maximum eigenvalue. A companion bound will be derived hereafter for the stochastic gradient descent algorithm.

Forward-backward splitting is an efficient method for minimizing convex cost functions with sparse regularization. It was originally derived for offline learning but a generalization of this algorithm for stochastic optimization, the so-called FOBOS, was proposed in \cite{Duchi2009}. It consists of using a stochastic approximation for $\nabla J$ at each iteration. This online approach can be easily coupled with the KLMS algorithm but, for convenience of presentation, we shall now describe the offline setup based on problem \eqref{eq:problem.parametric}.

\subsection{Application to KLMS algorithm}

In order to automatically discard the irrelevant elements from the dictionary $\cp{D}$, let us consider the minimization problem \eqref{eq:problem.parametric} with the sparsity-promoting convex regularization function $\Omega(\cdot)$
\begin{equation}
	\label{eq:klms.fbs}
        \balpha^* = \mathop{\arg\min}_{\balpha\in\R^N}\left\{Q(\balpha)=||\bd-\bK\balpha||^2+\lambda\Omega(\balpha)\right\}
\end{equation}
where $\bK$ is the $(N \times N)$ Gram matrix with $(i,j)$-th entry $\kappa(\vu_i,\vu_j)$. Problem \eqref{eq:klms.fbs} is of the form \eqref{eq:general.fbs}, and can be solved with the forward-backward splitting method. Two regularization terms are considered.

Firstly, we suggest the use of the well-known $\ell_1$-norm function defined as $\Omega_1(\balpha)=\sum_m |\alpha(m)|$. This regularization function is often used for sparse regression and its proximity operator is separable. Its $m$-th entry can be expressed as
\begin{equation}
	\label{eq:prox_l1}
    	\left(\text{Prox}_{\lambda\eta\|\cdot\|_1}(\balpha)\right)\!(m)=\text{sign}\{\alpha(m)\}\max\{|\alpha(m)|-\lambda\eta,0\}
\end{equation}
It is called the soft thresholding operator. One major drawback is that it promotes biased prediction.

Secondly, we consider an adaptive $\ell_1$-norm function of the form $\Omega_a(\balpha)=\sum_m w_m|\alpha(m)|$ where
the $w_m$'s are weights to be dynamically adjusted. The proximity operator for this regularization function is defined by
\begin{equation}
   	\label{eq:prox_l1a}
    	\left(\text{Prox}_{\lambda\eta\Omega_a(\cdot)}(\balpha)\right)(m)\!=\!\text{sign}\{\alpha(m)\} \max\{|\alpha(m)|-\lambda\eta\,w_m,0\}
\end{equation}
This regularization function has been proven to be more consistent than the usual $\ell_1$-norm \cite{Zou2006}, and tends to reduce the bias induced by the latter. Weights are usually chosen as $w_m=1/(|\alpha_\text{opt}(m)|+\epsilon_\alpha)$, where $\balpha_\text{opt}$ is the least-square solution of the problem \eqref{eq:problem.parametric}, and $\epsilon_\alpha$ a small constant to prevent the denominator from vanishing \cite{Candes2008}. Since $\balpha_\text{opt}$ is not available in our online case, we chose $w_m=1/(|\alpha_{n-1}(m)|+\epsilon_\alpha)$ at each iteration $n$. This technique, also referred to as reweighted least-square, is performed at each iteration of the stochastic optimization process. Note that a similar regularization term was used in \cite{Chen2009} in order to approximate the $\ell_0$-norm.

The pseudocode for KLMS algorithm with sparsity-promoting regularization, called FOBOS-KLMS, is provided in Algorithm 1. It can be noticed that the proximity operator is applied after the gradient descent step. The trivial dictionary elements associated with null coefficients in vector $\balpha_n$ are eliminated. This approach reduces to the generic KLMS algorithm in the case $\lambda=0$.

\begin{algorithm}[tb]
\caption{FOBOS-KLMS.}
\label{alg:Framwork}
\begin{algorithmic}[1]
\STATE \textbf{Initialization}\\
Select the step size $\eta$, and the parameters of the kernel; \\
Insert $\kappa(\cdot,\vu_1)$ into the dictionary, $\balpha_1=0$.\\
\STATE \textbf{for} $n=1,2,\cdots,$ do

\STATE
\begin{adjustwidth}{0.3cm}{0cm}
\textbf{if} $\max_{m=1,\ldots,M}|\kappa(\cb{u}_{n},\cb{u}_{\omega_m})|>\mu_0$\\
\begin{adjustwidth}{0.4cm}{0cm}
Compute $\bkappa_{\omega,n}$ and $\hat\balpha_n$ using equation \eqref{eq:klms.perso.1};\\
\end{adjustwidth}
\end{adjustwidth}

\STATE
\begin{adjustwidth}{0.3cm}{0cm}
\textbf{elseif} $\max_{m=1,\ldots,M}|\kappa(\cb{u}_{n},\cb{u}_{\omega_m})|\leq\mu_0$\\
\begin{adjustwidth}{0.4cm}{0cm}
Incorporate $\kappa(\cdot,\vu_n)$ into the dictionary;\\
Compute $\bkappa_{\omega,n}$ and $\hat\balpha_n$ using equation \eqref{eq:klms.perso.2};\\
\end{adjustwidth}
\end{adjustwidth}
\STATE
\begin{adjustwidth}{0.3cm}{0cm}
\textbf{end if}
\end{adjustwidth}
\STATE
\begin{adjustwidth}{0.3cm}{0cm}

$\balpha_{n}=\text{Prox}_{\lambda\eta\Omega(\cdot)}(\hat\balpha_n)$ using \eqref{eq:prox_l1} or \eqref{eq:prox_l1a}; \\
\end{adjustwidth}

\STATE \begin{adjustwidth}{0.3cm}{0cm}
Remove  $\kappa(\cdot,\vu_{\omega_m})$ from the dictionary if $\alpha_{n}(m)=0$.
\end{adjustwidth}

\STATE
 \begin{adjustwidth}{0.3cm}{0cm}
 The solution is given as \\
 $\psi(\vu_n)=\sum_{m=1}^{M}\alpha_m \kappa(\vu_n,\vu_{\omega_m})$.
 \end{adjustwidth}

\STATE \textbf{end for}

\end{algorithmic}
\end{algorithm}

\subsection{Stability in the mean}
\label{subsec:Stability-KLMS}

We shall now discuss the stability in mean of the FOBOS-KLMS algorithm. We observe that the KLMS algorithm with the sparsity inducing regularization can be written as
\begin{equation}
	\label{eq:weightupdate}
	 \balpha_n = \balpha_{n-1} + \eta \, e_n \,\bkappa_{\omega,n} - \bfun_{n-1}
\end{equation}
with
\begin{equation}
	\label{eq:ffun}
	f_{n-1}(m)=
	\begin{cases}
	\begin{split}
    	&\lambda\eta \, \text{sign}(\hat{\balpha}_{n-1}(m)) \qquad \text{if} \quad |\hat{\balpha}_{n-1}(m)| \geq\lambda\eta \\
	& \hat{\balpha}_{n-1}(m) \qquad\qquad\quad  \text{otherwise}
	\end{split}
	\end{cases}
\end{equation}
where $\hat{\balpha}_n = \balpha_{n-1} + \eta \, e_n \, \bkappa_{\omega,n}$. The function $\text{sign}(\alpha)$ is defined by
\begin{equation}
	\label{eq:funsign}
	\text{sign}(\alpha)=
	\begin{cases}
	\alpha/|\alpha| \qquad \alpha \neq 0 \\
	0 \qquad \qquad \text{otherwise}.
	\end{cases}
\end{equation}
Up to a variable change in $\lambda$, the general form \eqref{eq:weightupdate}-\eqref{eq:ffun} remains the same with the regularization function \eqref{eq:prox_l1a}. Note that the sequence $|f_{n-1}(m)|$ is bounded, by $\lambda\eta$ for the operator ~\eqref{eq:prox_l1}, and by $\lambda\eta/\epsilon_{\alpha}$ for the operator \eqref{eq:prox_l1a}.

\begin{theorem}
Assume MIA holds. For any initial condition~$\balpha_0$, the KLMS algorithm with sparsity promoting regularization \eqref{eq:prox_l1} and \eqref{eq:prox_l1a} asymptotically converges in the mean sense if the step-size $\eta$ is chosen to satisfy
\begin{equation}
	\label{eq:upper.bound2}
	0 < \eta < 2/\lambda_{\text{max}}(\mRkk)
\end{equation}
where $\mRkk=E\{\vk\vk^{\top}\}$ is the $(M \times M)$ correlation matrix of the kernelized input $\vk$.
\end{theorem}

To prove this theorem, we observe that the recursion \eqref{eq:error_vector1} for the weight error vector $\vv$ becomes
\begin{equation}
	\label{eq:weight_error1}
	 \vv = \bv_{n-1} - \eta \,\vk(\vk\,\bv_{n-1}+e^o_n) - \bfun_{n-1}
\end{equation}
Taking the expected value of both sides, and using the same assumptions as for \eqref{eq:stab_mean}, leads to
\begin{equation}
	\label{eq:weight_error2}
	 E\{\vv\} = (\bI - \eta\,\mRkk)^nE\{\bv_0\} + 
	 		\sum_{i=0}^{n-1}(\bI - \eta\,\mRkk)^iE\{\bfun_{n-i-1}\}
\end{equation}
with $\bv_0$ the initial condition. To prove the convergence of $E\{\vv\}$, we have to show that both terms on the r.h.s. converge as $n$ goes to infinity. The first term converges to zero if we can ensure that $\nu\triangleq\|\bI -\eta\,\mRkk\|<1$. We can easily check that this condition is met for any step-size $\eta$ satisfying the condition \eqref{eq:upper.bound2} since
\begin{equation}
	\label{eq:delta}
	\nu=|1 - \eta\,\lambda_{\text{max}}(\mRkk)|
\end{equation}
Let us show now that condition \eqref{eq:upper.bound2} also implies that the second term on the r.h.s. of equation \eqref{eq:weight_error2} asymptotically converges to a finite value, thus leading to the overall convergence of this recursion. First it has been noticed that the sequence $|f_{n-1}(m)|$ is bounded. Thus, each term of this series is bounded because
\begin{equation}
	\label{eq:seriesbound1}
	\begin{split}
	\|(\bI - \eta\,\mRkk)^i~E\{\bfun_{n-i-1}\}\| &\leq \|(\bI - \eta\,\mRkk)^i\|
			~E\{\|\bfun_{n-i-1}\|\} \\
		&\leq \sqrt{M}\,\nu^i\,f_{\text{max}}
	\end{split}
\end{equation}
where $f_{\text{max}}=\lambda\eta$ or $\lambda\eta/\epsilon_{\alpha}$, depending if one uses the regularization function \eqref{eq:prox_l1} or \eqref{eq:prox_l1a}. Condition \eqref{eq:upper.bound2} implies that $\nu<1$ and, as a consequence,
\begin{equation}
	\label{eq:seriesbound2}
	\sum_{i=0}^{n-1}\|(\bI - \eta\,\mRkk)^i~E\{\bfun_{n-i-1}\}\| \leq \frac{\sqrt{M}\,f_{\text{max}}}{1-\nu}
\end{equation}
The second term on the r.h.s. of equation \eqref{eq:weight_error2} is an absolutely convergent series. This implies that it is a convergent series. Because the two terms of equation \eqref{eq:weight_error2} are convergent series, we finally conclude that $E\{\vv\}$ converges to a steady-state value if condition \eqref{eq:upper.bound2} is satisfied. Before concluding this section, it should be noticed that we have shown in \cite{Parreira2012} that
\begin{equation}
	\label{eq:maxeigenvalue}
	\lambda_{\text{max}}(\mRkk) = \rmd + (M-1)\,\rod.
\end{equation}
Parameters $\rmd$ and $\rod$ are given by expression \eqref{eq:Rkk_mdf} in the case of a possibly partially matching dictionary.

\subsection{Simulation Results of Proposed Algorithm}
\label{sec:AlgorithmSimulation}

We shall now illustrate the good performance of the FOBOS-KLMS algorithm with the two examples considered in Section \ref{sec:BeahviorAnalysis}. Experimental settings were unchanged, and the results were averaged over $200$ Monte Carlo runs. The coherence threshold $\mu_0$ in Algorithm \ref{alg:Framwork} was set to $0.01$.

\begin{figure*}[!htb]
\centering
\subfigure[MSE]
{	\includegraphics[width=6cm]{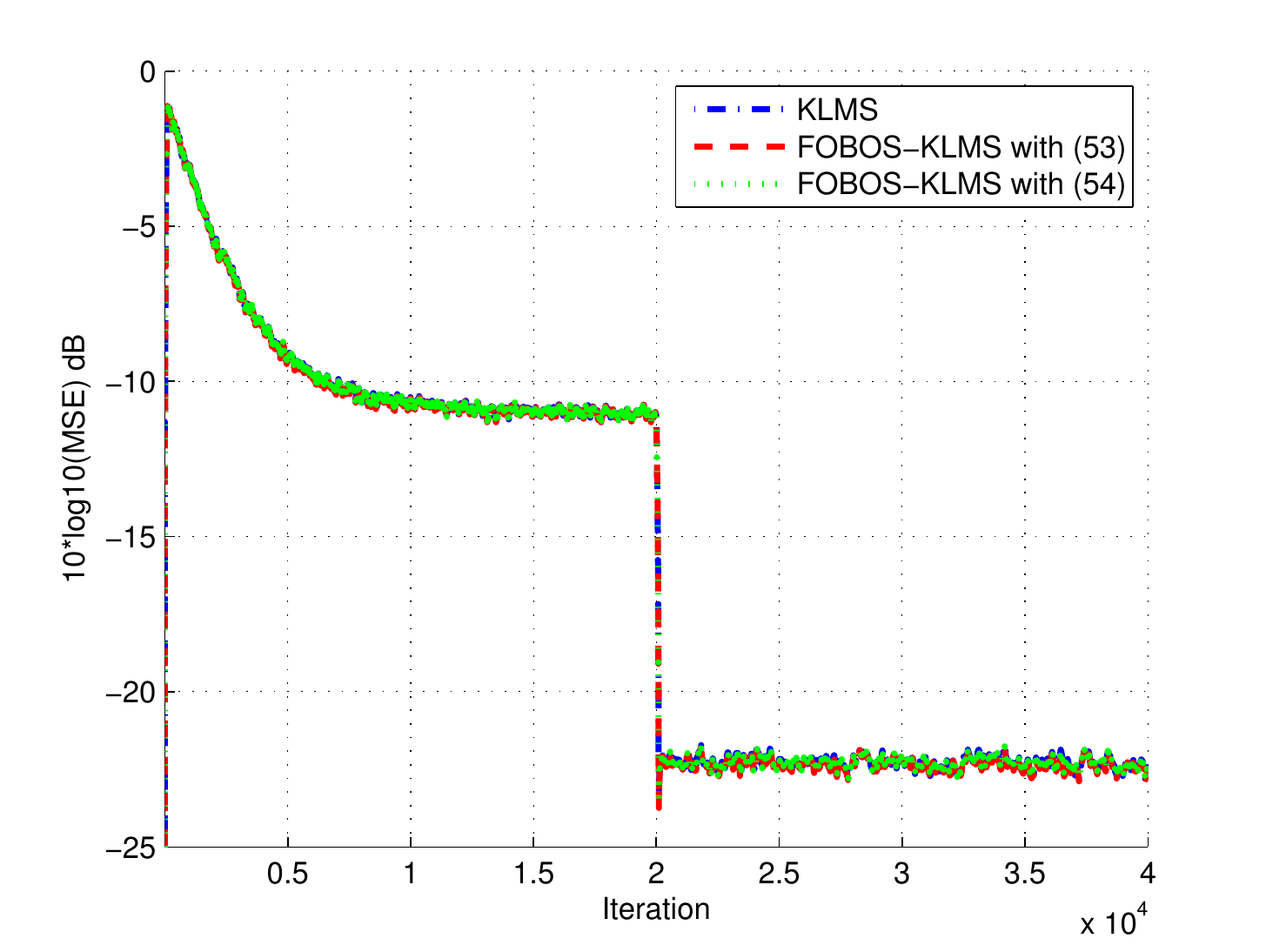}}
\subfigure[Evolution of the size of dictionary ]
{\includegraphics[width=6cm]{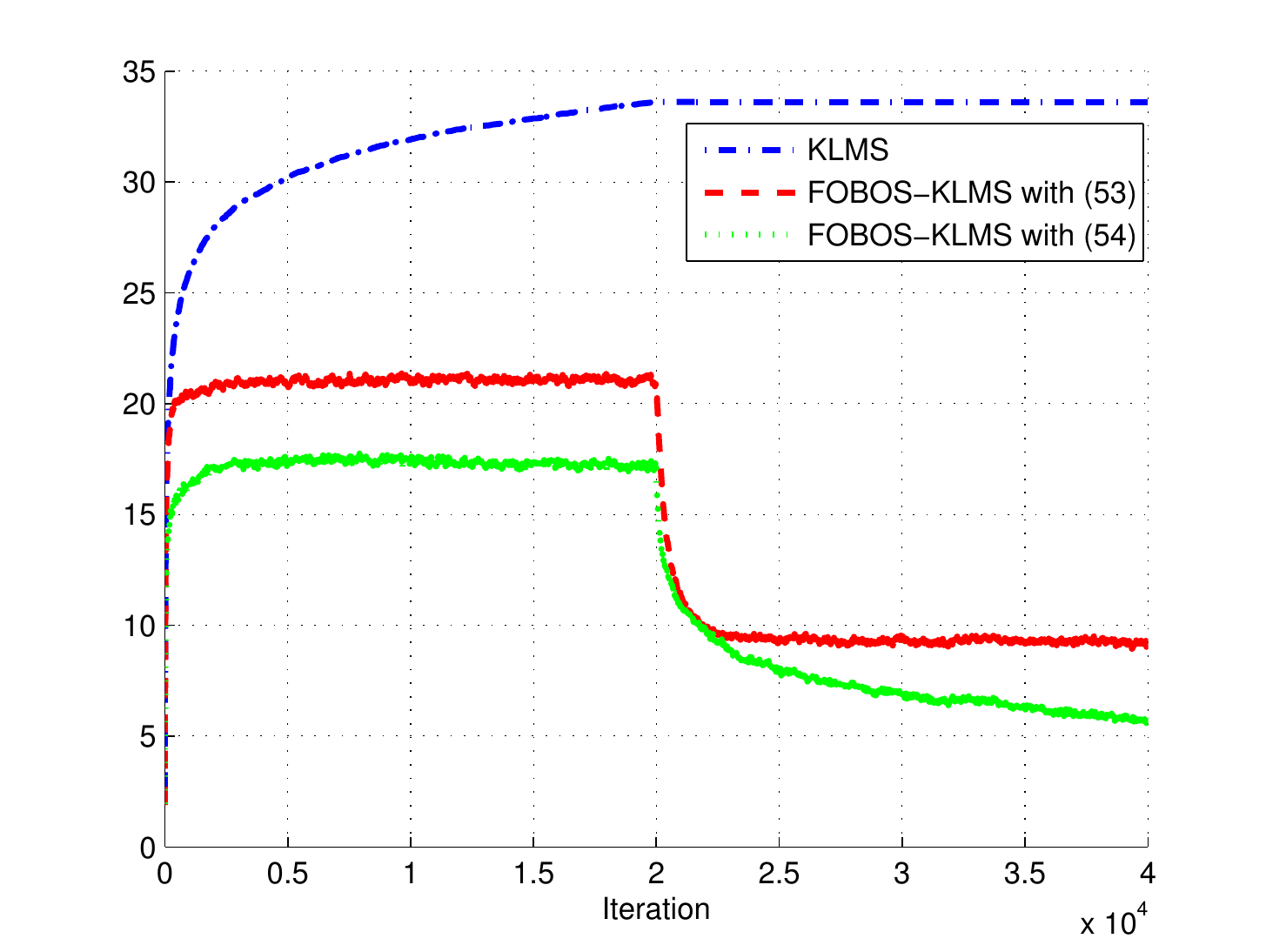}}
\caption{Learning curves for Example 1 where $\sigma_u:\,0.35 \rightarrow 0.15$.}
\label{fig:Exa1_1}
\end{figure*}

\begin{figure*}[!htb]
\centering
\subfigure[MSE]
{	\includegraphics[width=6cm]{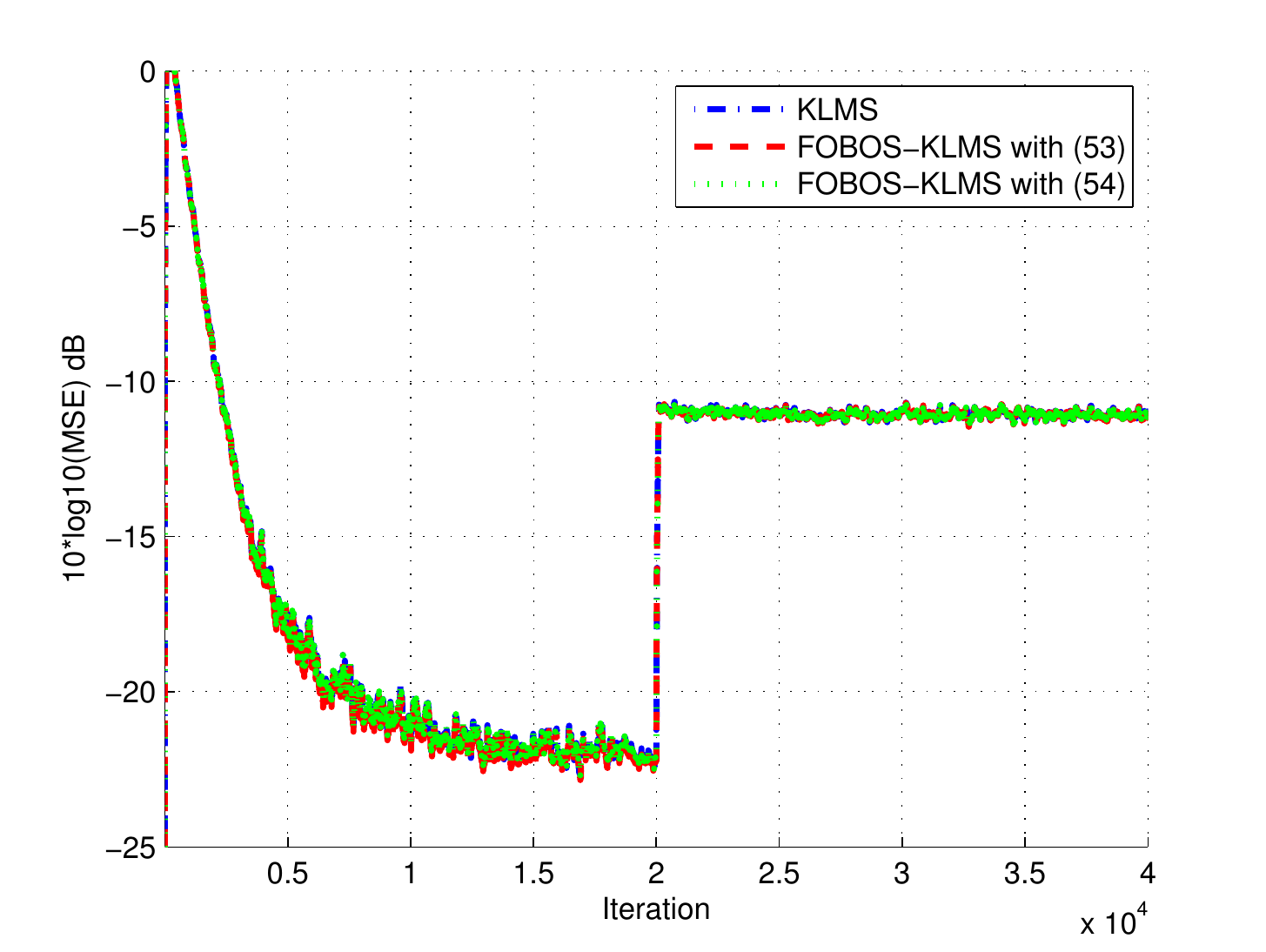}}
\subfigure[Evolution of the size of dictionary ]
{\includegraphics[width=6cm]{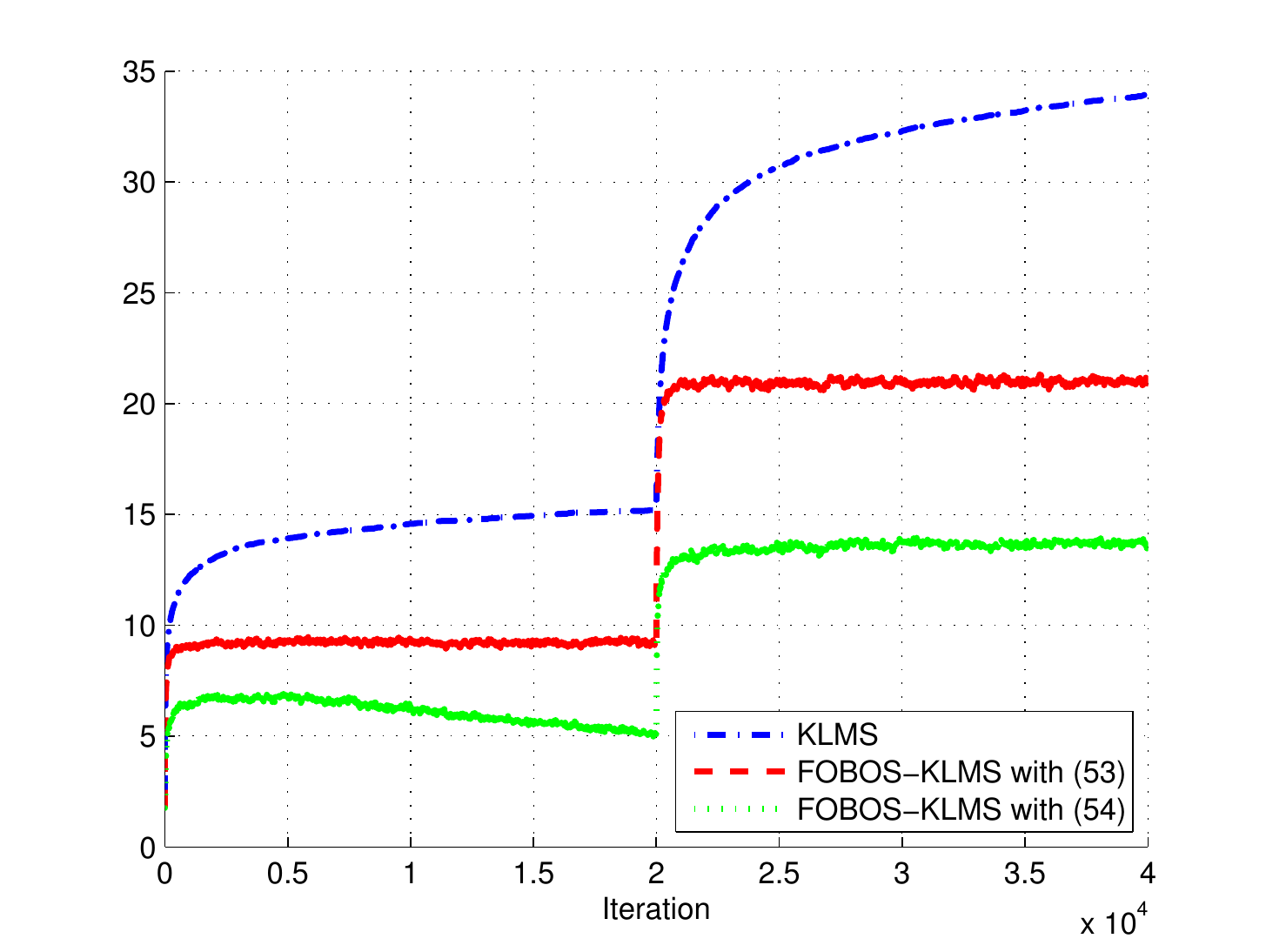}}
\caption{Learning curves for Example 1 where $\sigma_u:\,0.15 \rightarrow 0.35$.}
\label{fig:Exa1_2}
\end{figure*}

\begin{figure*}[!htb]
\centering
\subfigure[MSE]
{	\includegraphics[width=6cm]{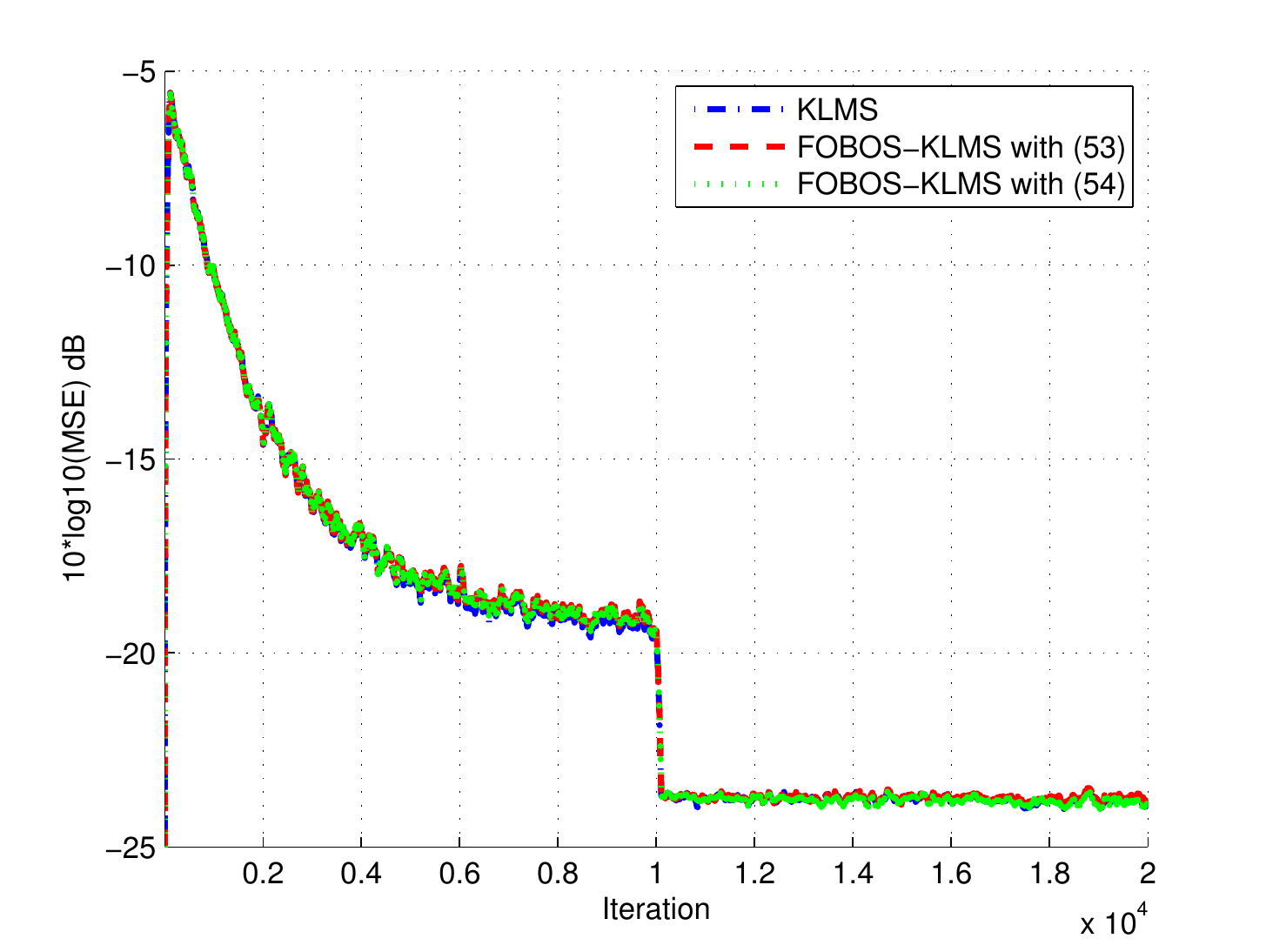}}
\subfigure[Evolution of the size of dictionary ]
{\includegraphics[width=6cm]{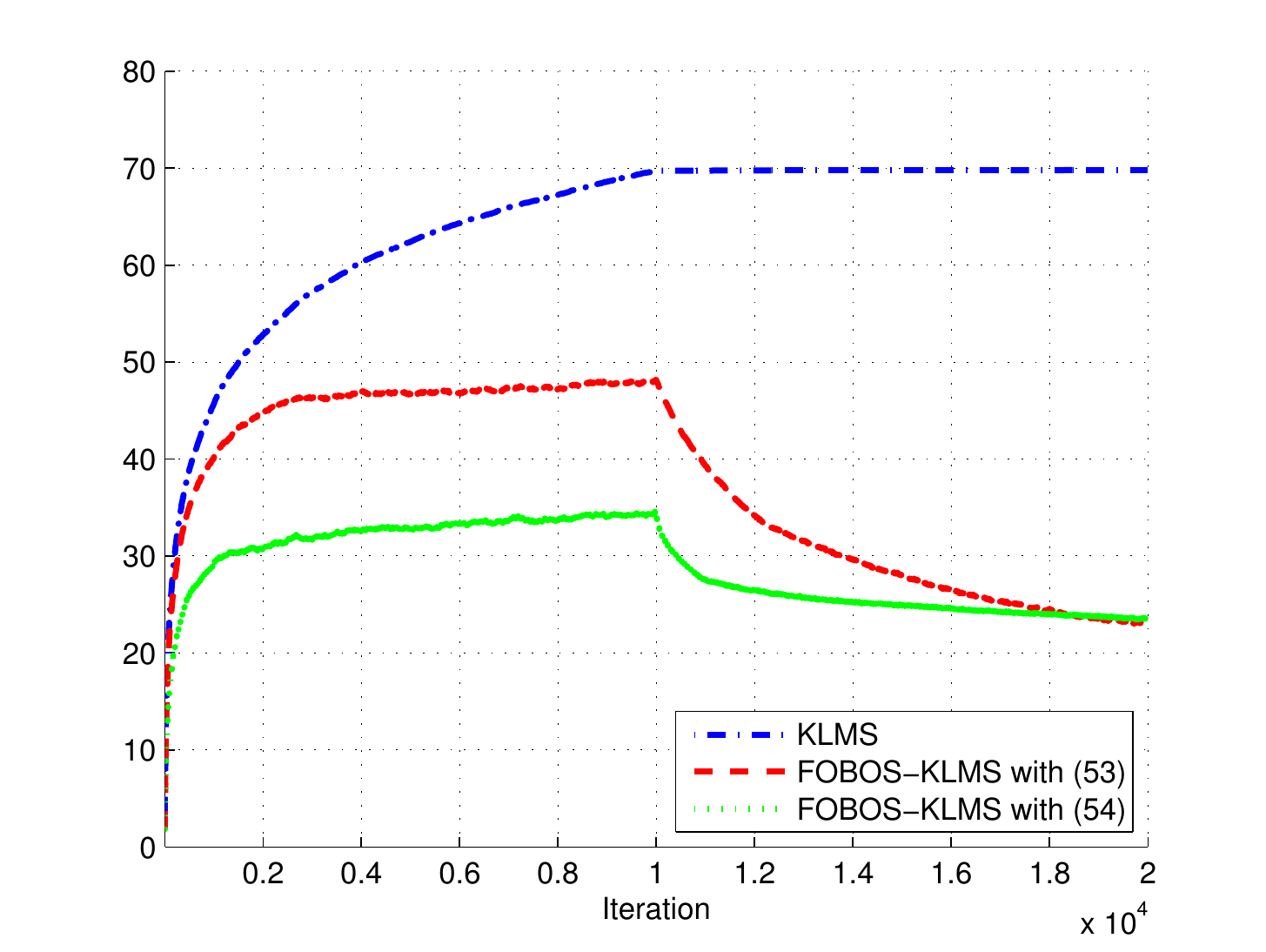}}
\caption{Learning curves for Example 2 with $\sigma_{u_2},\sigma_{v_u}:\,$ \protect\\ $\sqrt{0.0656} \rightarrow \sqrt{0.0156}$.}
\label{fig:Expa2_1}
\end{figure*}

\begin{figure*}[!htb]
\centering
\subfigure[MSE]
{	\includegraphics[width=6cm]{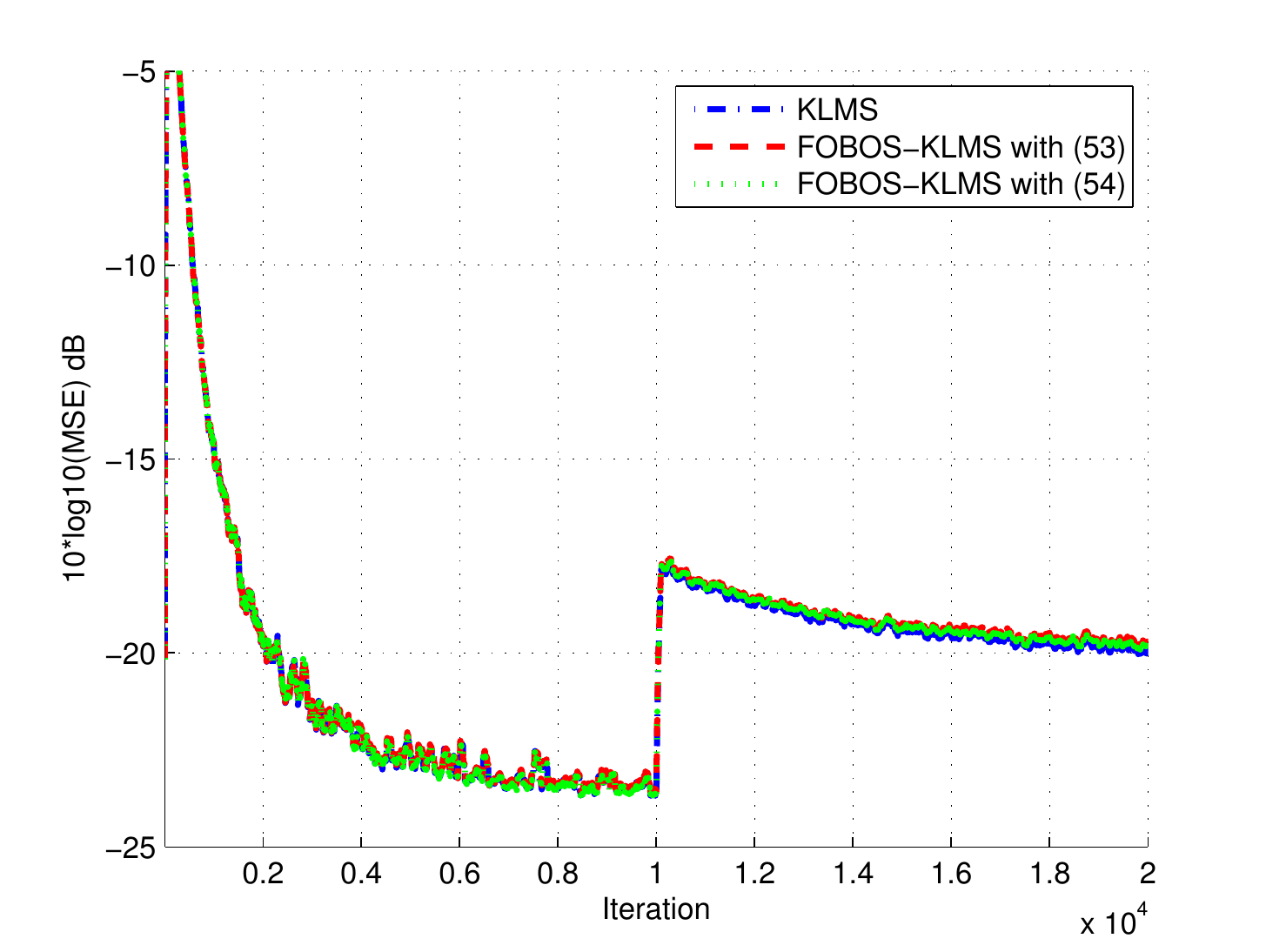}}
\subfigure[Evolution of the size of dictionary ]
{\includegraphics[width=6cm]{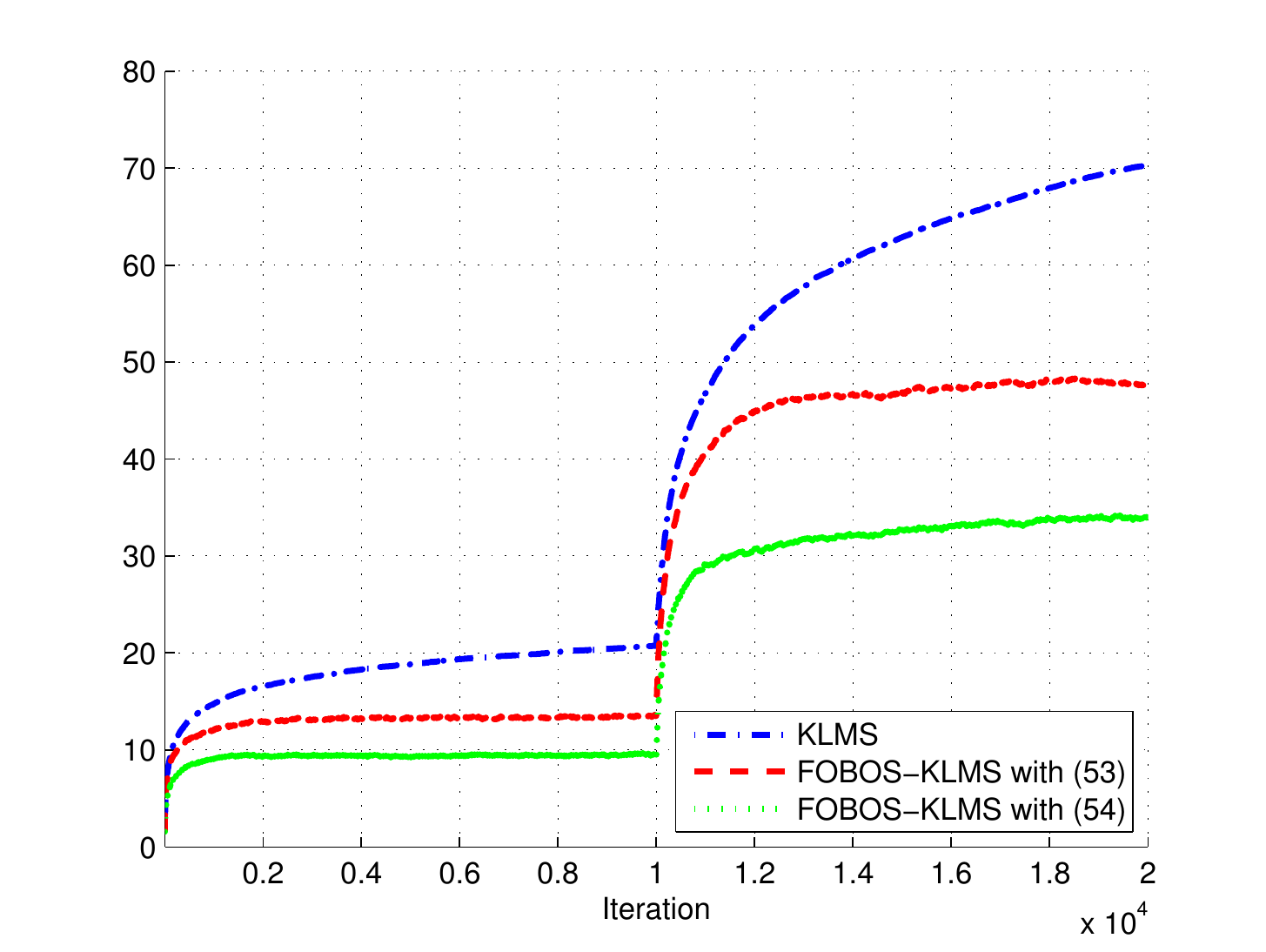}}
\caption{Learning curves for Example 2 with $\sigma_{u_2},\sigma_{v_u}:\,$ \protect\\ $\sqrt{0.0156} \rightarrow \sqrt{0.0656}$}
\label{fig:Expa2_2}
\end{figure*}

One can observe in Figures \ref{fig:Exa1_2} and \ref{fig:Expa2_2} that the size of the dictionary designed by the KLMS with coherence criterion dramatically increases when the variance of the input signal increases. In this case, this increased dynamic forces the algorithm to pave the input space $\cp{U}$ with additional dictionary elements. In Figures \ref{fig:Exa1_1} and \ref{fig:Expa2_1}, the algorithm does not face this problem since the variance of the input signal abruptly decreases. The dictionary update with new elements is suddenly stopped. Again, these two scenarios clearly show the need for dynamically updating the dictionary by adding or discarding elements. Figures \ref{fig:Exa1_1} to \ref{fig:Expa2_2} clearly illustrate the merits of the FOBOS-KLMS algorithm with the regularizations \eqref{eq:prox_l1} and \eqref{eq:prox_l1a}. Both principles efficiently control the structure of the dictionary as a function of instantaneous characteristics of the input signal. They  significantly reduce the order of the KLMS filter without affecting its performance.

\section{Conclusion}
\label{sec:conclusion}

In this paper, we presented an analytical study of the convergence behavior of the Gaussian least-mean-square algorithm in the case where the statistics of the dictionary elements only partially match the statistics of the input data. This allowed us to emphasize the need for updating the dictionary in an online way, by discarding the obsolete elements and adding appropriate ones. We introduced the so-called FOBOS-KLMS algorithm, based on forward-backward splitting to deal with $\ell_1$-norm regularization, in order to automatically adapt the dictionary to the instantaneous characteristics of the input signal. The stability in the mean of this method was analyzed, and a condition on the step-size for convergence was derived. The merits of FOBOS-KLMS were illustrated by simulation examples.

\bibliographystyle{IEEEtran}
\bibliography{bibliography}

\end{document}